\RequirePackage{fix-cm}
\documentclass[twocolumn,compsoc]{CVM}

\usepackage{amsmath,amssymb} 
\usepackage{amsfonts}
\usepackage{times}
\usepackage{soul}
\usepackage{cite}
\usepackage{epsfig}
\usepackage{cite}
\usepackage{multirow}
\usepackage{diagbox}
\usepackage{rotating}
\usepackage{overpic}
\usepackage{array}
\usepackage{epsfig}
\usepackage{colortbl}
\usepackage{etoolbox}
\usepackage{footnote}
\usepackage{pifont}

\newcommand{\xmark}{\text{\ding{55}}}
\newcommand{\cmark}{\ding{51}}%

\definecolor{gray1}{rgb}{.8,.8,.8}

\newcommand{\PreserveBackslash}[1]{\let\temp=\\#1\let\\=\temp}
\newcolumntype{C}[1]{>{\PreserveBackslash\centering}p{#1}}
\newcolumntype{R}[1]{>{\PreserveBackslash\raggedleft}p{#1}}
\newcolumntype{L}[1]{>{\PreserveBackslash\raggedright}p{#1}}

\def\ie{\emph{i.e.,~}}
\def\eg{\emph{e.g.,~}}
\def\etal{{\em et al.~}}

\definecolor{gray1}{rgb}{.8,.8,.8}

\newcommand{\myPara}[1]{\vspace{.05in}\noindent\textbf{#1}}

\graphicspath{{./figs/}}

\renewcommand{\arraystretch}{1.1}
\renewcommand{\tabcolsep}{.5mm}

\newcommand{\AddImg}[1]{}
\newcommand{\AddImgs}[2]{}
\newcommand{\AddImgsU}[2]{}

\usepackage[pagebackref=false,letterpaper=true,colorlinks=true,linkcolor=black,citecolor=black,bookmarks=false]{hyperref}

\graphicspath{{figs/},{figures/},{figures/cvm/}}

\def\ManuscriptInfo{Manuscript received: 201x-xx-xx; accepted: 201x-xx-xx.}

\def\PaperTitle{Salient Object Detection: A Survey}

\def\AuthorNames{Ali~Borji$^1$,
        Ming-Ming~Cheng$^2$ \rm{\Letter},
        Qibin Hou$^2$,
        Huaizu Jiang$^3$,
        Jia Li$^4$}

\def\AuthorAdress{$1\quad $ & MarkableAI. E-mail: ali@markable.ai.\\
    $2\quad $ & TKLNDST, College of Computer Science, Nankai University. E-mail: cmm@nankai.edu.cn \\
    $3\quad $ & University of Massachusetts Amherst. \\
    $4\quad $ & Beihang University. Email: jiali@buaa.edu.cn. \\
}

\def\firstheader{DOI 10.1007/s41095-019-0149-9\hfill Vol. 5, No. 2, June 2019, 117--150\\[5mm]~Article Type (Research/Review)}

\headevenname{{Ali Borji \etal} }

\begin{document}

\MakePageStyle

\MakeAbstract{
Detecting and segmenting salient objects in natural scenes, often referred to as salient object detection, has attracted a lot of interest in computer vision. While many models have been proposed and several applications have emerged, yet a deep understanding of achievements and issues is lacking. We aim to provide a comprehensive review of the recent progress in salient object detection and situate this field among other closely related areas such as generic scene segmentation, object proposal generation, and saliency for fixation prediction. Covering 228 publications, we survey i) roots, key concepts, and tasks, ii) core techniques and main modeling trends, and iii) datasets and evaluation metrics in salient object detection. We also discuss open problems such as evaluation metrics and dataset bias in model performance and suggest future research directions.
}

\MakeKeywords{Salient object detection, bottom-up saliency, explicit saliency, visual attention, regions of interest}

\section{Introduction}


Humans are able to detect visually distinctive, so called salient, scene regions effortlessly and rapidly (\ie~pre-attentive stage).
These filtered regions are then perceived and processed in
finer details for the extraction of
richer high-level information (\ie~attentive stage).
This capability has long been studied by cognitive scientists and has recently attracted
a lot of interest in the computer vision community mainly because it helps
find the objects or regions that efficiently represent a scene and
thus harness complex vision problems such as scene understanding.
Some topics that are closely or remotely related to visual saliency include:
salient object detection~\cite{cheng2015global}, fixation prediction~\cite{bylinskii2015saliency,bylinskii2016should},
object importance~\cite{spain2011measuring,berg2012understanding,m2013fixations},
memorability~\cite{isola2011makes}, scene clutter~\cite{rosenholtz2007measuring},
video interestingness~\cite{katti2008pre,gygli2013interestingness,dhar2011high,jiang2013understanding},
surprise~\cite{itti2005bayesian}, image quality assessment~\cite{wang2004image,wang2002image,zhang2016application},
scene typicality~\cite{vogel2004semantic,ehinger2011estimating}, aesthetic~\cite{dhar2011high}
and attributes~\cite{farhadi2009describing}.
Given space limitations, this paper cannot fully explore all the aforementioned research directions.
Instead, we only focus on salient object detection, a research area that has been greatly developed in the past twenty years in particular since 2007~\cite{liu2007region}.



\subsection{What is Salient Object Detection about?}
\label{whatIS}

``Salient object detection'' or ``salient object segmentation''
is commonly interpreted in computer vision as a process that
includes two stages: \textit{
1) detecting the most salient object and
2) segmenting the accurate region of that object}.
Rarely, however,  models explicitly distinguish between these
two stages (with few exceptions such as
\cite{mishra2012active,liXiaodiCVPR2014,borjiTIP2014}).
Following the seminal works by Itti \etal \cite{itti1998model}
and Liu \etal\cite{LiuSZTS07Learn},
models adopt the saliency concept to simultaneously perform
the two stages together.
This is witnessed by the fact that these stages have not
been separately evaluated.
Further, mostly area-based scores have been employed for
model evaluation (\eg~Precision-recall).
The first stage does not necessarily need to be limited to only one object.
The majority of existing models, however,
attempt to segment the most salient object,
although their prediction maps can be used to
find several objects in the scene.
The second stage falls in the realm of classic segmentation problems
in computer vision but with the difference that
here accuracy is only determined by the most salient object.


\begin{figure}[t]
    \centering  
    \includegraphics[width=8.3cm,height=3cm]{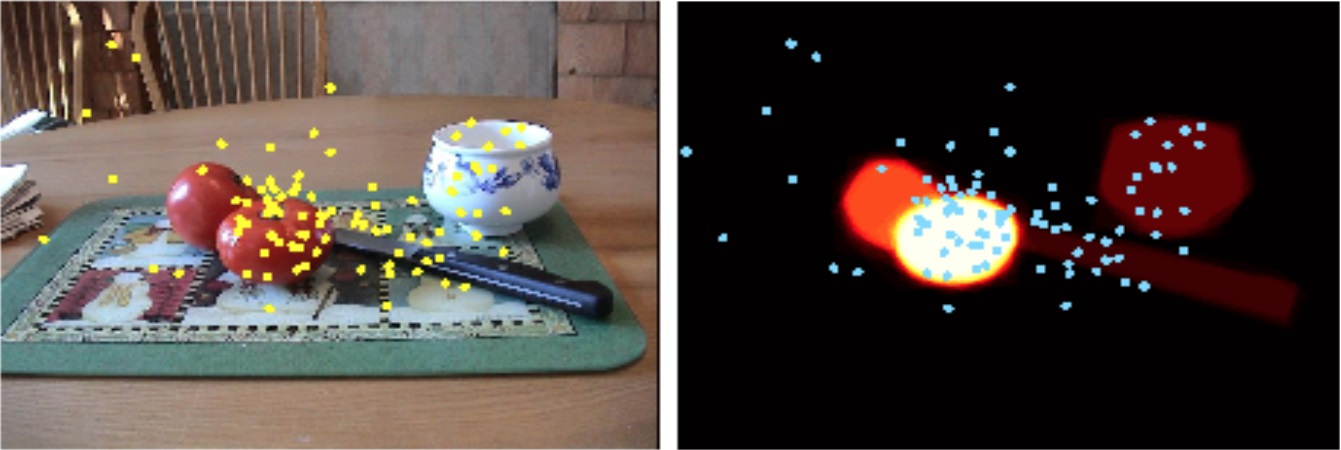} \\
    \vspace{-5pt}
    \caption{An example image in Borji~\etal's experiment
       \cite{borji2013stands} along with annotated salient objects.
       Dots represent 3-second free-viewing fixations.
    }\label{fig:salObjExp}
    \vspace{-15pt}
\end{figure}

\begin{figure*}[t]
    \centering  
    \includegraphics[width=\linewidth]{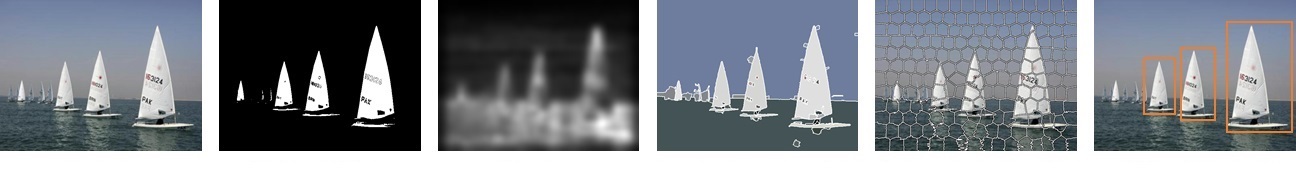}
        \vspace{-25pt}
    \caption{Sample results produced by different models. From left to right: input image, salient object detection \cite{perazzi2012saliency}, fixation prediction \cite{itti1998model}, image segmentation (regions with various sizes)~\cite{ComaniciuM02MeanShift}, image segmentation (superpixels with comparable sizes)~\cite{AchantaSSLFS12slic}, and object proposals (true positives) \cite{BingObj2014}.
    }\label{fig:researchdiff}

    \vspace{-15pt}
\end{figure*}


In general, it is agreed that for good saliency detection a model
should meet at least the following three criteria:
1) \textit{good detection:} the probability of missing real salient
regions and falsely marking the background as a salient region should be low,
2) \textit{high resolution:} saliency maps should have high or full resolution
to accurately locate salient objects and retain original image information,
and 3) \textit{computational efficiency:} as front-ends to other
complex processes, these models should detect salient regions quickly.



\subsection{Situating Salient Object Detection}

%

Salient object detection models usually aim to detect
only the most salient objects in a scene and
segment the whole extent of those objects.
Fixation prediction models, on the other hand, typically try to predict where humans look,
\ie~a small set of fixation points~\cite{borji2013state,borji2013analysis}.
Since the two types of methods output a single continuous-valued saliency map,
where a higher value in this map indicates that the corresponding image
pixel is more likely to be attended, they can be used interchangeably.

A strong correlation exists between fixation locations and salient objects. Further, humans often agree which each other when asked to choose the most salient object in a scene~\cite{borji2013stands,borjiTIP2014,liXiaodiCVPR2014}. These are illustrated in \figref{fig:salObjExp}.

Unlike salient object detection and fixation prediction models,
object proposal models aim at producing
a small set, typically a few hundreds or thousands, of
overlapping candidate object bounding boxes or region proposals~\cite{hosang2016makes}.
Object proposal generation and salient object detection
are highly related. Saliency estimation is explicitly used as a cue in objectness
methods \cite{alexe2010object,siva2013looking}.

%
%
%

Image segmentation, a.k.a semantic scene labeling or semantic segmentation,
is one of the very well researched areas in computer vision
(\eg \cite{cheng2001color}).
In contrast to salient object detection where the output is a binary map, these models aim to assign a label, one out of several classes such as sky, road, and building, to each image pixel.

%


Fig.~\ref{fig:researchdiff} illustrates the difference among these research themes.

\begin{figure*}[t]
  \vspace*{-20pt}
  \centering  
  \begin{overpic}[width=.95\linewidth]{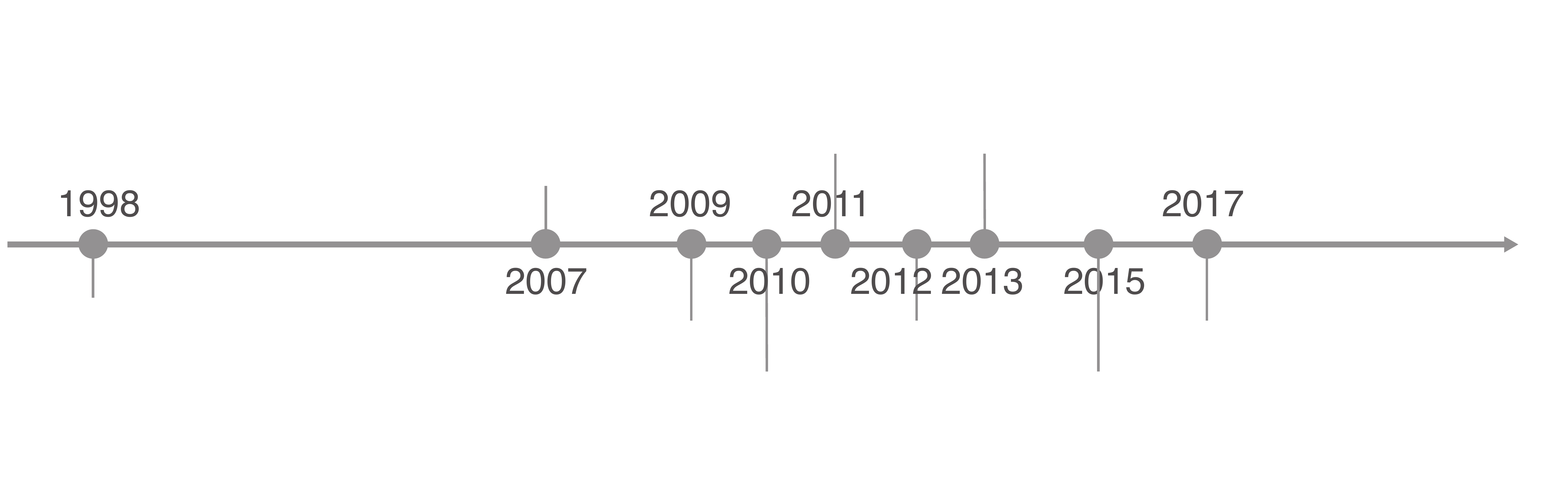}
    \put(1,10){\textbf{First wave}: Itti \etal\cite{itti1998model},}
    \put(1,8){computational model}
    \put(14,24){\textbf{Second wave}: Liu \etal \cite{LiuSZTS07Learn},}
    \put(14,22){def. as binary labeling prob.,}
    \put(14,20){dataset with bound. boxes}
    \put(25,9){Achanta \etal \cite{achanta2009frequency}: pixel}
    \put(25,7){accuracy g-truth dataset}
    \put(42,24){Cheng \etal \cite{ChengCVPR11}:}
    \put(42,22){global contrast}
    \put(42,4){Goferman \etal \cite{goferman2012context}:}
    \put(42,2){context aware saliency}
    \put(53,9){Perazzi \etal \cite{perazzi2012saliency}:}
    \put(53,7){saliency filters}
    \put(60,24){New models \& datasets:}
    \put(60,22){\cite{JiangWYWZL13,margolin2013saliency,yan2013hierarchical,yang2013graph}}
    \put(67,5){\textbf{Third wave}: deep models}
    \put(67,3){\cite{SuperCNN_IJCV2015,wang2015Deep,zhao2015saliency,li2015visual,zou2015harf}}
    \put(78,12){Hou \etal \cite{hou2016deeply}:}
    \put(78,10){deeply supervised}
  \end{overpic}
   \vspace*{-10pt}
  \caption{A simplified chronicle of salient object detection modeling.
      The first wave started with the Itti \etal~model~\cite{itti1998model}
      followed by the second hit with the introduction of the Liu \etal
      \cite{LiuSZTS07Learn} who were the first to define saliency
      as a binary segmentation problem.
      The third wave started with the surge of deep learning models and the
      Li \etal~model~\cite{li2015visual}. }\label{fig:coronicle}
\vspace{-10pt}
\end{figure*}

\subsection{History of Salient Object Detection}

One of the earliest saliency models, proposed by Itti \etal \cite{itti1998model}, generated the \emph{first wave} of interest across multiple
disciplines including cognitive psychology, neuroscience, and
computer vision.
This model is an implementation of earlier general computational
frameworks and psychological theories of bottom-up attention based on
center-surround mechanisms (\eg~\textit{Feature Integration Theory}
by Treisman and Gelade~\cite{treisman1980feature},
\textit{Guided Search Model} by Wolfe~\etal~\cite{wolfe1989guided},
and the \textit{Computational Attention Architecture} by Koch and Ullman~\cite{koch1987shifts}).
In~\cite{itti1998model}, Itti~\etal~show some examples where their model is able to
detect spatial discontinuities in scenes.
Subsequent behavioral (\eg~\cite{parkhurst2002modeling})
and computational investigations (\eg~\cite{bruce2005saliency})
used fixations as a means to verify the saliency hypothesis and to compare models. 

A \emph{second wave} of interest surged with the works of Liu
\etal~\cite{LiuSZTS07Learn,liu2011learning} and Achanta
\etal~\cite{achanta2008salient}
who defined saliency detection as a binary segmentation problem.
These authors were inspired by some earlier models striving to detect salient regions or proto-objects
(\eg~Ma and Zhang~\cite{ma2003contrast}, Liu and Gleicher~\cite{LiuG06Region},
and Walther~\etal~\cite{walther2006modeling}).
A plethora of saliency models has emerged since then.
%
It has been, however, less clear how this new definition relates to other established computer vision areas such as
image segmentation  (\eg~\cite{arbelaez2011contour,martin2004learning}),
category independent object proposal generation (\eg~\cite{alexe2010object,endres2010category,BingObj2014}),
fixation prediction (\eg~\cite{bruce2005saliency,judd2009learning,hou2007saliency,borji2012exploiting,borji2012boosting}),
and object detection (\eg~\cite{viola2001rapid,felzenszwalb2010object}).

A third wave of interest has appeared recently with the resurgence of the convolutional
neural networks (CNNs) \cite{lecun1998gradient}, in particular with the
introduction of the fully convolutional neural networks \cite{long2015fully}.
Unlike the majority of classic methods based on contrast cues \cite{cheng2015global},
CNN-based methods eliminate the need for hand-crafted features,
alleviate the dependency on center bias knowledge,
and hence have been adopted by many researchers.
A CNN-based model normally contains hundreds of thousands of
tunable parameters and neurons with variable receptive field sizes.
Neurons with large receptive fields provide global information that
can help better identify the most salient region in an image.
While neurons with small receptive fields provide local information
that can be leveraged to refine saliency maps produced by the
top layers.
This allows highlighting salient regions and refining their boundaries.
%
These desirable properties enable CNN-based models to achieve
unprecedented performance
compared to hand-crafted feature-based models. CNN models are gradually
becoming the mainstream direction in salient object detection.

\section{Survey of the State of the Art}
\label{sec:survey}

In this section, we review related works in 3 categories, including: 1) salient object detection models, 2) applications , and 3) datasets. Due to similarity among some models, such that it is sometimes hard to draw sharp boundaries among them, here we mainly focus on the models contributing to the major ``waves'' in the chronicle shown in Fig.~\ref{fig:coronicle}.



\subsection{Old Testament: Classic Models}\label{sec:sodm}

A large number of approaches have been proposed for detecting salient objects in images in the past two decades.
Except for a few models which attempt to segment objects-of-interest (\eg~\cite{hua2006iterative,ko2006automatic,allili2007object}),
most of these approaches aim to identify the salient subsets\footnote{
Visual subsets could be pixels, blocks, superpixels and regions.
Blocks are rectangular patches uniformly sampled from the image (pixels are $1\times1$ blocks).
A superpixel or a region is perceptually homogeneous image patch that is confined with intensity edges.
Superpixels, in the same image, often have comparable but different sizes,
while the shapes and sizes of regions may change remarkably.}
from images first (\ie~compute a saliency map) and then integrate them
to segment the entire salient object.


In general, classic approaches can be categorized in two different ways depending on the types operation or attributes they exploit.

\noindent \textbf{1)~Block-based vs. Region-based analysis}.
Two types of visual subsets have been utilized: blocks and regions\footnote{In this review, the term ``block'' is used to represent pixels and patches,
while ``superpixel'' and ``region'' are used interchangeably.},
to detect salient objects.
Blocks were primarily adopted by early approaches, while regions became popular with the introduction of superpixel algorithms.

\noindent \textbf{2)~Intrinsic cues vs. Extrinsic cues}.
A key step in detecting salient objects is to distinguish them from distractors.
To this end, some approaches propose to extract various cues only from the input image itself
to highlight targets and to suppress distractors (\ie~the intrinsic cues).
However, other approaches argue that intrinsic cues are often insufficient to distinguish
targets and distractors specially when they share common visual attributes.
To overcome this issue, they incorporate extrinsic cues such as user annotations,
depth map, or statistical information of similar images to facilitate detecting salient objects
in the image.

From the above model categorization, four combinations are thus possible. For a better organization, we group the models in 
three major subgroups \textit{1) block-based models with intrinsic cues}, \textit{2) region-based models with intrinsic cues},
and \textit{3) models with extrinsic cues (both block- and region-based)}.
Some approaches that may not easily fit into these subgroups will be discussed under the \textit{other classic models} subgroup. Reviewed models are listed in~\tabref{tab:salientObjModelsIntrin} (Intrinsic models),~\tabref{tab:SalModelExtrinsic} (Extrinsic models), and~\tabref{tab:salModelsOthers} (Other classic models).

\subsubsection{Block-based Models with Intrinsic Cues}\label{sec:groupEI}

\newcommand{\calN}{{\mathcal N}}
In this subsection, we mainly review salient object detection models
which utilize intrinsic cues extracted from blocks.
Following the seminal work of Itti~\etal~\cite{itti1998model},
salient object detection is widely defined as capturing the uniqueness, distinctiveness,
or rarity in a scene.

In early works~\cite{ma2003contrast,LiuG06Region,achanta2008salient},
uniqueness was often computed as the pixel-wise center-surround contrast.
Hu \etal~\cite{hu2005robust} represent the input image in a 2D space
using the polar transformation of its features. Each region in the image is then mapped
into a 1D linear subspace.
Afterwards, the Generalized Principal Component Analysis (GPCA)~\cite{vidal2005generalized}
is used to estimate the linear subspaces without actually segmenting the image.
Finally, salient regions are selected by measuring feature contrasts
and geometric properties of regions.
Rosin~\cite{rosin2009simple} proposes an efficient approach for detecting salient objects.
His approach is parameter-free and requires only very simple pixel-wise operations
such as edge detection, threshold decomposition and moment preserving binarization.
%
Valenti~\etal~\cite{valenti2009image} propose an isophote-based framework
where the saliency map is estimated by linearly combining the saliency maps computed
in terms of curvedness, color boosting, and isocenters clustering.

In an influential study, Achanta~\etal~\cite{achanta2009frequency}
adopt a frequency-tuned approach to compute full resolution saliency maps.
The saliency of pixel $x$ is computed as:
\begin{equation}
  s(x)=\|I_{\mu}-I_{\omega_{hc}}(x)\|^2,
\end{equation}
where $I_{\mu}$ is the mean pixel value of the image (\eg~RGB/Lab features)
and $I_{\omega_{hc}}$ is a Gaussian blurred version of the input image
(\eg~using a $5\times5$ kernel).

Without any prior knowledge of the sizes of salient objects,
multi-scale contrast is frequently adopted for robustness purposes \cite{LiuG06Region, LiuSZTS07Learn}.
A $L$-layer Gaussian pyramid is first constructed (as in
\cite{LiuG06Region, LiuSZTS07Learn}).
The saliency score of pixel $x$ at the image at the $l$th-level of this pyramid (denoted as $I^{(l)}$)
is defined as:
\begin{align}
s(x) = \sum_{l=1}^L\sum_{x'\in\calN(x)} ||I^{(l)}(x) - I^{(l)}(x')||^2,
\end{align}
where $\calN(x)$ is a neighboring window centered at $x$ (e.g., $9\times9$ pixels).
Even with such multi-scale enhancement,
intrinsic cues derived at pixel-level are often too poor to support object segmentation.
To address this, some works (e.g.,~\cite{LiuSZTS07Learn, achanta2008salient,klein2011center,LiLSDH13Contextual}) extended the contrast analysis to the patch level (i.e., a patch compared to its neighbors). 

Later in~\cite{klein2011center}, Klein and Frintrop proposed an information-theoretic approach to compute center-surround contrasts using the Kullback-Leibler divergence between distribution of features such as intensity, color and orientation.
Li~\etal~\cite{LiLSDH13Contextual} formulated the center-surround contrast as
a cost-sensitive max-margin classification problem.
The center patch is labeled as a positive sample
while the surrounding patches are all used as negative samples.
The saliency of the center patch is then determined by its separability from
surrounding patches based on a trained cost-sensitive Support Vector Machine (SVM).

Some works have defined patch uniqueness as its global contrast with
other patches~\cite{goferman2012context}.
Intuitively, a patch is considered to be salient if it is remarkably distinct from its most similar patches,
while their spatial distances are taken into account.
Similarly, Borji and Itti computed local and global patch rarity in RGB and LAB color spaces and fused them to
predict fixation locations~\cite{borji2012exploiting}.
In a recent work~\cite{margolinmakes},
Margolin~\etal~propose to define the uniqueness of a patch
by measuring its distance to the average patch based on the observation
that distinct patches are more scattered than non-distinct ones in the high-dimensional space.
To further incorporate the patch distributions,
the uniqueness of a patch is measured by projecting its path to the average patch
onto the principal components of the image.


To sum up, approaches in \secref{sec:groupEI} aim to detect
salient objects based on pixels or patches
where only intrinsic cues are utilized.
These approaches usually suffer from two shortcomings:
i) high-contrast edges usually stand out instead of the salient object,
and ii) the boundary of the salient object is not preserved well (especially when using large blocks).
To overcome these issues, some methods propose to compute saliency based on regions.
This offers two main advantages.
First, the number of regions is far less than the number of blocks,
which implies the potential to develop highly efficient and fast algorithms.
Second, more informative features can be extracted from regions,
leading to better performance.
These region-based approaches will be discussed in the next subsection.



\subsubsection{Region-based Models with Intrinsic Cues}\label{sec:groupLI}

Saliency models in the second subgroup adopt intrinsic cues extracted
from image \emph{regions} generated using methods such as graph-based segmentation
\cite{felzenszwalb2004efficient},
mean-shift~\cite{ComaniciuM02MeanShift},
SLIC~\cite{AchantaSSLFS12slic} or Turbopixels~\cite{LevinshteinSKFDS09Turbop}.
Different from the block-based models,
region-based models often segment an input image into regions aligned with
intensity edges first and then compute a regional saliency map.

As an early attempt, in~\cite{LiuG06Region},
the regional saliency score is defined as the average saliency score of
its contained pixels,
defined in terms of multi-scale contrast.
Yu~\etal~\cite{yu2007rule} propose a set of rules to determine the background scores
of each region based on observations from background and salient regions.
Saliency, defined as uniqueness in terms of \textbf{global regional contrast},
is widely studied in many approaches
\cite{ChengPAMI,yan2013hierarchical,scharfenberger2013statistical,ChengWLZVC13Efficient,JiangD13submodular}.
In~\cite{ChengPAMI}, a region-based saliency algorithm is introduced by measuring
the global contrast between the target region with respect to all other image regions.
In a nutshell, an image is first segmented into $N$ regions $\{r_i\}_{i=1}^N$.
Saliency of the region $r_i$ is measured as:
\begin{equation}
s(r_i)=\sum_{j=1}^N w_{ij}D_r(r_i,r_j),
\label{eq:globalContrast}
\end{equation}
where $D_r(r_i,r_j)$ captures the appearance contrast between two regions.
Higher saliency scores are assigned to regions with large global contrast.
$w_{ij}$ is a weight term between regions $r_i$ and $r_j$,
which incorporates spatial distance and region size.
%
%
%
%
Perazzi \etal~\cite{perazzi2012saliency}
demonstrate that if $D_r(r_i,r_j)$ is defined as the Euclidean distance of colors
between $r_i$ and $r_j$,
the global contrast can be computed using 
efficient filtering based techniques \cite{adams2010fast}.

In addition to color uniqueness, distinctiveness of complementary cues
such as texture~\cite{scharfenberger2013statistical} and
structure~\cite{shi2013pisa} is also considered for salient object detection.
Margolin~\etal~\cite{margolinmakes} propose to combine the
regional uniqueness and patch distinctiveness to form a saliency map. 
Instead of maintaining a hard region index for each pixel,
a soft abstraction is proposed in~\cite{ChengWLZVC13Efficient} to generate
a set of large scale perceptually homogeneous regions using histogram
quantization and Gaussian Mixture Models (GMM).
By avoiding the hard decision boundaries of superpixels,
such soft abstraction provides large spatial support which results in a more uniform saliency region.

\begin{figure*}
  \centering
  \small
  \renewcommand{\arraystretch}{.8}
  \renewcommand{\tabcolsep}{3.3mm}
  \begin{tabular}{|c||l|cc|c|c|c|c|c|}
  \hline
  \multirow{2}{*}{\#} & \multirow{2}{*}{\textbf{Model}} & \multirow{2}{*}{\textbf{Pub}} & \multirow{2}{*}{\textbf{Year}} & \multirow{2}{*}{\textbf{Elements}} & \multicolumn{2}{c|}{\textbf{Hypothesis}} & \textbf{Aggregation} & \multirow{2}{*}{\textbf{Code}}  \\
  \cline{6-7}
   & & & & & \textbf{Uniqueness} & \textbf{Prior} & (\textbf{Optimization}) &  \\
  \hline \hline
  1 & \textbf{FG}~\cite{ma2003contrast} & MM & 2003 & PI & L & - & - & NA \\
  2 & \textbf{RSA}~\cite{hu2005robust} & MM & 2005 & PA & G & - & - & NA\\
  3 & \textbf{RE}~\cite{LiuG06Region} & ICME & 2006 & mPI + RE & L & - & LN & NA  \\
  4 & \textbf{RU}~\cite{yu2007rule} & TMM & 2007 & RE & - & P & LN & NA \\
  5 & \textbf{AC}~\cite{achanta2008salient} & ICVS & 2008 & mPA & L & - & LN & NA  \\
  6 & \textbf{FT}~\cite{achanta2009frequency} & CVPR & 2009 & PI & CS & - & - & C \\
  7 & \textbf{ICC}~\cite{valenti2009image} & ICCV & 2009 & PI & L & - & LN & NA  \\
  8 & \textbf{EDS}~\cite{rosin2009simple} & PR & 2009 & PI & - & ED & - & NA   \\
  9 & \textbf{CSM}~\cite{yu2010automatic} & MM & 2010 & PI + PA & L & SD & - & NA  \\
  10 & \textbf{RC}~\cite{ChengPAMI} & CVPR & 2011 & RE & G & - & - & C \\
  11 & \textbf{HC}~\cite{ChengPAMI} & CVPR & 2011 & RE & G & - & - & C \\
  12 & \textbf{CC}~\cite{lu2011salient} & ICCV & 2011 & mRE & - & CV & - & NA  \\
  13 & \textbf{CSD}~\cite{klein2011center} & ICCV & 2011 & mPA & CS & - & LN & NA  \\
  14 & \textbf{SVO}~\cite{chang2011fusing} & ICCV & 2011 & PA + RE & CS & O & EM & M + C \\
  15 & \textbf{CB}~\cite{jiang2011automatic} & BMVC & 2011 & mRE & L & CP & LN & M + C \\
  16 & \textbf{SF}~\cite{perazzi2012saliency} & CVPR & 2012 & RE & G & SD & NL & C \\
  17 & \textbf{ULR}~\cite{shen2012unified} & CVPR & 2012 & RE & SPS & CP + CLP & - & M + C  \\
  18 & \textbf{GS}~\cite{wei2012geodesic} & ECCV & 2012 & PA/RE & - & B & - & NA \\
  19 & \textbf{LMLC}~\cite{XieEtAlTIP2013} & TIP & 2013 & RE & CS & - & BA & M + C \\
  20 & \textbf{HS}~\cite{yan2013hierarchical} & CVPR & 2013 & hRE & G & - & HI & EXE \\
  21 & \textbf{GMR}~\cite{YangZLRY13Manifold} & CVPR & 2013 & RE & - & B & - & M \\
  22 & \textbf{PISA}~\cite{shi2013pisa} & CVPR & 2013 & RE & G & SD + CP & NL & NA  \\
  23 & \textbf{STD}~\cite{scharfenberger2013statistical} & CVPR & 2013 & RE & G & - & - & NA  \\
  24 & \textbf{PCA}~\cite{margolinmakes} & CVPR & 2013 & PA + PE & G & - & NL & M+C  \\
  25 & \textbf{GU}~\cite{ChengWLZVC13Efficient} & ICCV & 2013 & RE & G & - & - & C \\
  26 & \textbf{GC}~\cite{ChengWLZVC13Efficient} & ICCV & 2013 & RE & G & SD & AD & C \\
  27 & \textbf{CHM}~\cite{LiLSDH13Contextual} & ICCV & 2013 & PA + mRE & CS + L & - & LN & M + C \\
  28 & \textbf{DSR}~\cite{li2013saliency} & ICCV & 2013 & mRE & - & B & BA & M + C \\
  29 & \textbf{MC}~\cite{Jiang2013Saliency} & ICCV & 2013 & RE & - & B & - & M + C \\
  30 & \textbf{UFO}~\cite{JiangLYP13UFO} & ICCV & 2013 & RE & G & F + O & NL & M + C \\
  31 & \textbf{CIO}~\cite{jia2013category} & ICCV & 2013 & RE & G & O & GMRF & NA  \\
  32 & \textbf{SLMR}~\cite{zou2013segmentation} & BMVC & 2013 & RE & SPS & BC & - & NA \\
  33 & \textbf{LSMD}~\cite{peng2013salient} & AAAI & 2013 & RE & SPS & CP + CLP & - & NA  \\
  34 & \textbf{SUB}~\cite{JiangD13submodular} & CVPR & 2013 & RE & G & CP + CLP + SD & - & NA \\
  35 & \textbf{PDE}~\cite{liu2014adaptive} & CVPR & 2014 & RE & - & CP + B + CLP & - & NA \\
  36 & \textbf{RBD}~\cite{zhu2014saliency} & CVPR & 2014 & RE & - & BC & LS & M  \\
  \hline
  \end{tabular}
  \caption{Salient object detection models with intrinsic cues (sorted by year).
    Element, \{PI = pixel, PA = patch, RE = region\}, where prefixes m and h indicate multi-scale and hierarchical versions, respectively.
    Hypothesis, \{CP = center prior, G = global contrast, L = local contrast,
    ED = edge density, B = background prior, F = focusness prior, O = objectness prior,
    CV = convexity prior, CS = center-surround contrast, CLP = color prior,
    SD = spatial distribution, BC = boundary connectivity prior, SPS = sparse noises\}.
    Aggregation/optimization, \{LN = linear, NL = non-linear, AD = adaptive, HI = hierarchical, BA = Bayesian, GMRF = Gaussian MRF, EM = energy minimization, and LS = least-square solver.\}.
    Code, \{M= Matlab, C= C/C++, NA = not available, EXE = executable\}.
  }\label{tab:salientObjModelsIntrin}
  \vspace{-10pt}
\end{figure*}

In \cite{jiang2011automatic}, Jiang \etal~propose a \textbf{multi-scale local region contrast}
based approach, which calculates saliency values across multiple segmentations for robustness purposes and
combines these regional saliency values to obtain a pixel-wise saliency map.
%
%
%
A similar idea for estimating regional saliency using multiple hierarchical segmentations
is adopted in~\cite{yan2013hierarchical,li2013saliency}.
%
Li~\etal~\cite{LiLSDH13Contextual} extend the pairwise local contrast by building a hypergraph,
constructed by non-parametric multi-scale clustering of superpixels,
to capture both internal consistency and external separation of regions.
Salient object detection is then casted as finding salient vertices and hyperedges in the hypergraph.


\newcommand{\calG}{{\mathcal G}}
\newcommand{\calV}{{\mathcal V}}
\newcommand{\calE}{{\mathcal E}}

\newcommand{\vF}{{\textbf F}}
\newcommand{\vS}{{\textbf S}}
\newcommand{\vL}{{\textbf L}}
\newcommand{\vf}{{\textbf f}}
\newcommand{\vl}{{\textbf l}}
\newcommand{\vs}{{\textbf s}}
\newcommand{\calR}{{\mathcal R}}
Salient objects, in terms of uniqueness,
can also be defined as the \textbf{sparse noises} in a certain feature space
where the input image is represented as a low-rank matrix
\cite{shen2012unified,zou2013segmentation,peng2013salient}.
The basic assumption is that non-salient regions (\ie~background) can be explained by the low-rank matrix
while the salient regions are indicated by the sparse noises.
%
%

Based on such a general low-rank matrix recovery framework,
Shen and Wu~\cite{shen2012unified} propose a unified approach to incorporate traditional
low-level features with higher-level guidance, \eg~\textbf{center prior}, \textbf{face prior}, and \textbf{color prior},
to detect salient objects based on a learned feature transformation\footnote{Though extrinsic ground-truth annotations are adopted to
learn high-level priors and the feature transformation,
we classify this model in intrinsic models to better organize the low-rank matrix recovery based approaches.
Additionally, we treat face and color priors as universal intrinsic cues
for salient object detection.}.
Instead, Zou~\etal~\cite{zou2013segmentation} propose to exploit bottom-up segmentation
as a guidance cue of the low-rank matrix recovery for robustness purpose.
Similar to~\cite{shen2012unified}, high-level priors are also adopted in~\cite{peng2013salient},
where a tree-structured sparsity-inducing norm regularization is introduced to hierarchically
describe the image structure in order to uniformly highlight the entire salient object.

\newcommand{\vP}{{\textbf P}}

In addition to capturing the uniqueness,
more and more priors are proposed for salient object detection as well.
\textbf{Spatial distribution prior}~\cite{LiuSZTS07Learn} implies that the wider a color is
distributed in the image, the less likely a salient object contains this color.
%
The spatial distribution of superpixels can be efficiently evaluated
in linear time using the Gaussian blurring kernel as well,
in a similar way of computing the global regional contrast in
\eqnref{eq:globalContrast}.
Such a spatial distribution prior is also considered in~\cite{shi2013pisa}
evaluated in terms of both color and structure cues.

Center prior assumes that a salient object is more likely to be found near the image center.
In other words, the background tends to be far away from the image center.
To this end, the \textbf{backgroundness prior} is adopted for salient object detection
\cite{wei2012geodesic,YangZLRY13Manifold, li2013saliency, Jiang2013Saliency},
assuming that a narrow border of the image is the background region, \ie~the pseudo-background.
With this pseudo-background as a reference, regional saliency can be computed
as the contrast of regions versus ``background''.
%
%
In~\cite{YangZLRY13Manifold}, a two-stage saliency computation framework is proposed based on
the manifold ranking on an undirected weighted graph.
In the first stage, the regional saliency scores are computed based on the relevances
given to each side of the pseudo-background queries.
In the second stage, the saliency scores are refined based on the relevances given to the initial foreground.
%
%
In~\cite{li2013saliency}, saliency computation is formulated as
the dense and sparse reconstruction errors w.r.t. the pseudo-background.
The dense reconstruction error of each region is computed based on the Principal Component Analysis
(PCA) basis of the background templates,
while the sparse reconstruction error is defined as the residual based on the sparse representation
of the background templates.
These two types of reconstruction errors are propagated to pixels on multiple segmentations,
which will be fused to form the final saliency map.
%
Jiang~\etal~\cite{Jiang2013Saliency} propose to formulate
the saliency detection via absorbing Markov Chain
where the transient and absorbing nodes are superpixels around
the image center and border, respectively.
The saliency of each superpixel is computed as the absorbed time
for the transient node to the absorbing nodes of the Markov Chain.




Beyond these approaches, the generic \textbf{objectness prior}\footnote{Although it is learned from training data,
we also tend to treat it as
a universal intrinsic cue for salient object detection.}
is also used to facilitate salient object detection by leveraging object proposals~\cite{alexe2010object}.
Chang~\etal~\cite{chang2011fusing} present a computational
framework by fusing the
objectness and regional saliency into a graphical model.
These two terms are jointly estimated by iteratively minimizing
the energy function that encodes their mutual interactions.
In~\cite{JiangLYP13UFO}, regional objectness is defined as the
average objectness values of its contained pixels,
which will be incorporated for regional saliency computation.
Jia and Han~\cite{jia2013category} compute the saliency of
each region by comparing it to the ``soft'' foreground
and background according to the objectness prior.
%


Salient object detection relying on the pseudo-background
assumption may fail sometimes,
especially when the object touches the image border.
To this end, a \textbf{boundary connectivity prior}
is utilized in \cite{ChengPAMI,zhu2014saliency}.
Intuitively, salient objects are much less connected to
the image border than the ones in the background.
Thus, the boundary connectivity score of a region
could be estimated according to the ratio
between its length along the image border and the spanning
area of this region~\cite{zhu2014saliency},
which can be computed based on its geodesic distances to the
pseudo-background and other regions, respectively.
%
Such a boundary connectivity score is then integrated into a
quadratic objective function
to get the final optimized saliency map.
It is worth noting that similar ideas of boundary connectivity
prior are also investigated in~\cite{zou2013segmentation}
as \emph{segmentation prior} and
as \emph{surroundness} in~\cite{zhang2013boolean}.

The \textbf{focusness prior},
the fact that a salient object is often photographed in focus
to attract more attention,
has been investigated in~\cite{JiangLYP13UFO, li2014saliency}.
Jiang~\etal~\cite{JiangLYP13UFO} calculate the focusness from
the degree of focal blur.
By modeling such a de-focus blur as the convolution of a sharp
image with a point spread function,
approximated by a Gaussian kernel,
the pixel-level focusness is casted as estimating the
standard deviation of the Gaussian kernel by scale space analysis.
Regional focusness score is computed by propagating the focusness
and/or sharpness at the boundary and interior edge pixels.
The saliency score is finally derived from the non-linear combination
of uniqueness (global contrast), objectness, and focusness scores.



Performance of salient object detection based on regions might be affected by the segmentation parameters.
In addition to other approaches based on multi-scale regions
\cite{jiang2011automatic, yan2013hierarchical, LiLSDH13Contextual},
single-scale potential salient regions are extracted by solving the facility location problem
in~\cite{JiangD13submodular}.
An input image is first represented as an undirected graph on superpixels,
where a much smaller set of candidate region centers are then generated through agglomerative clustering.
On this set, a submodular objective function is built to maximize the similarity.
By applying a greedy algorithm, the objective function can be iteratively
optimized to group superpixels into regions
whose saliency values are further measured via the regional global contrast and spatial distribution.

The Bayesian framework is exploited for saliency computation
\cite{rahtu2010segmenting,XieEtAlTIP2013}, formulated as estimating the posterior probability of pixel $x$ being foreground given the input image $I$.
%
To estimate the saliency prior, a convex hull $H$ is first
estimated around the detected interest points.
The convex hull $H$, which divides the image $I$ into the
inner region $R_I$ and outside region $R_O$,
provides a coarse estimation of foreground as well as
background and can be adopted for likelihood computation.
%
Liu~\etal~\cite{liu2014adaptive} adopt an optimization-based framework for detecting salient objects. Similar to~\cite{XieEtAlTIP2013},
a convex hull is roughly estimated 
to partition an image into pure background and potential foreground.
Then, saliency seeds are learned from the image,
while a guidance map is learned from background regions, as well as human prior knowledge.
Using these cues,
a general Linear Elliptic System with Dirichlet boundary is introduced to
model the diffusions from seeds to other regions to generate a saliency map.

Among all the models reviewed in this subsection, there are mainly three types of regions adopted for saliency computation. Irregular regions with varying sizes can be generated using the graph-based segmentation algorithm~\cite{felzenszwalb2004efficient}, mean-shift algorithm~\cite{ComaniciuM02MeanShift}, or clustering (quantization). On the other hand, with recent progress on superpixels algorithms, compact regions with comparable sizes are also popular choices using the SLIC algorithm~\cite{AchantaSSLFS12slic}, Turbopixel algorithm~\cite{LevinshteinSKFDS09Turbop}, etc. The main difference between these two types of regions is whether the influence of region size should be taken into account. Furthermore, soft regions are also considered for saliency analysis, where every pixel maintains a probability belonging to each of all the regions (components) instead of only a hard region label (\eg~fitted by a GMM). To further enhance robustness of segmentation, regions can be generated based on multiple segmentations or in a hierarchical way. Generally, single-scale segmentation is faster, while multi-scale segmentation can improve the overall performance.

To measure the saliency of regions, uniqueness, usually in the form of global and local regional contrasts, is still the most frequently used feature. In addition, more and more complementary priors for the regional saliency are investigated to improve the overall performance, such as backgroundness, objectness, focusness and boundary connectivity. Compared with the block-based saliency models, the extension of these priors is also the main advantage of the region-based saliency models. Furthermore, regions provide more sophisticated cues (\eg~color histogram) to better capture the salient object of a scene in contrast to pixels and patches. Another benefit of defining saliency using regions is related to the efficiency. Since the number of regions in an image is far less than the number of pixels, computing saliency at region level can significantly reduce the computational cost while producing full-resolution saliency maps.

Notice that the approaches discussed in this subsection only utilize intrinsic cues. In the next subsection, we will review how to incorporate extrinsic cues to facilitate the detection of salient objects.

\begin{figure*}
  \centering
  \small
   \renewcommand{\arraystretch}{.82}
  
  \begin{tabular}{|c||l|cc|c|c|c|c|c|c|c|}
    \hline
    \multirow{2}{*}{\#} & \multirow{2}{*}{\textbf{Model}} & \multirow{2}{*}{\textbf{Pub}} & \multirow{2}{*}{\textbf{Year}} & \multirow{2}{*}{\textbf{Cues}} & \multirow{2}{*}{\textbf{Elements}} & \multicolumn{2}{c|}{\textbf{Hypothesis}} & \textbf{Aggregation} & \multirow{2}{*}{\textbf{GT Form}} & \multirow{2}{*}{\textbf{Code}} \\
  \cline{7-8}
   & & & & & & \textbf{Uniqueness} & \textbf{Prior} & (\textbf{Optimization}) & &  \\
    \hline
    1 & \textbf{LTD}~\cite{LiuSZTS07Learn} & CVPR & 2007 & GT & mPI + PA + RE & L + CS & SD & CRF & BB & NA  \\
    2 & \textbf{OID}~\cite{khuwuthyakorn2010object} & ECCV & 2010 & GT & mPI + PA + RE & L + CS & SD & mixtureSVM & BB & NA \\
    3 & \textbf{LGCR}~\cite{mehrani2010saliency} & BMVC & 2010 & GT & RE & - & P & BDT & BM & NA \\
    4 & \textbf{DRFI}~\cite{JiangWYWZL13} & CVPR & 2013 & GT & mRE & L & B + P & RF & BM & M + C  \\
    5 & \textbf{LOS}~\cite{lu2014learning} & CVPR & 2014 & GT & RE & L + G & PRA + B + SD + CP & SVM & BM & NA \\
    6 & \textbf{HDCT}~\cite{kim2014salient} & CVPR & 2014 & GT & RE & L + G & SD + P + HD & BDT + LS & BM & M \\
    \hline \hline
    \multirow{2}{*}{\#} & \multirow{2}{*}{\textbf{Model}} & \multirow{2}{*}{\textbf{Pub}} & \multirow{2}{*}{\textbf{Year}} & \multirow{2}{*}{\textbf{Cues}} & \multirow{2}{*}{\textbf{Elements}} & \multicolumn{2}{c|}{\textbf{Hypothesis}} & \textbf{Aggregation} & \multirow{2}{*}{\textbf{GT Necessity}} & \multirow{2}{*}{\textbf{Code}}  \\
  \cline{7-8}
   & & & & & & \textbf{Uniqueness} & \textbf{Prior} & (\textbf{Optimization}) & &  \\
    \hline
    7 & \textbf{VSIT}~\cite{marchesotti2009framework} & ICCV & 2009 & SI & PA & - & SS & - & yes & NA \\
    8 & \textbf{FIEC}~\cite{wang2011image} & CVPR & 2011 & SI & PI + PA & L & - & LN & no & NA \\
    9 & \textbf{SA}~\cite{MaiNL13Aggregation} & CVPR & 2013 & SI & PI & - & CMP & CRF & yes & NA\\
    10 & \textbf{LBI}~\cite{siva2013looking} & CVPR & 2013 & SI & PA & SP & - & - & no & M + C \\
    \hline \hline
    \multirow{2}{*}{\#} & \multirow{2}{*}{\textbf{Model}} & \multirow{2}{*}{\textbf{Pub}} & \multirow{2}{*}{\textbf{Year}} & \multirow{2}{*}{\textbf{Cues}} & \multirow{2}{*}{\textbf{Elements}} & \multicolumn{2}{c|}{\textbf{Hypothesis}} & \textbf{Aggregation} & \multirow{2}{*}{\textbf{Type}} & \multirow{2}{*}{\textbf{Code}} \\
  \cline{7-8}
   & & & & & & \textbf{Uniqueness} & \textbf{Prior} & (\textbf{Optimization}) & &  \\
    \hline
    11 & \textbf{LC}~\cite{zhai2006visual} & MM & 2006 & TC & PI + PA & L & - & LN & online & NA  \\
    12 & \textbf{VA}~\cite{liu2008video} & ICPR & 2008 & TC & mPI + PA + RE & L & CS + SD + MCO & CRF & offline & NA \\
    13 & \textbf{SEG}~\cite{rahtu2010segmenting} & ECCV & 2010 & TC & PA + PI & CS & MCO & CRF & offline & M + C  \\
    14 & \textbf{RDC}~\cite{bin2013temporally} & CSVT & 2013 & TC & RE & L & - & - & offline & NA  \\
    \hline \hline
    \multirow{2}{*}{\#} & \multirow{2}{*}{\textbf{Model}} & \multirow{2}{*}{\textbf{Pub}} & \multirow{2}{*}{\textbf{Year}} & \multirow{2}{*}{\textbf{Cues}} & \multirow{2}{*}{\textbf{Elements}} & \multicolumn{2}{c|}{\textbf{Hypothesis}} & \textbf{Aggregation} & \multirow{2}{*}{\textbf{Image Number}} & \multirow{2}{*}{\textbf{Code}} \\
    \cline{7-8}
   & & & & & & \textbf{Uniqueness} & \textbf{Prior} & (\textbf{Optimization}) & &  \\
    \hline
    15 & \textbf{CSIP}~\cite{li2011co} & TIP & 2011 & SCO & mRE & - & RS & LN & two & M + C  \\
    16 & \textbf{CO}~\cite{ChangLL11from} & CVPR & 2011 & SCO & PI + PA & G & RP & - & multiple & NA \\
    17 & \textbf{CBCO}~\cite{fu2013cluster} & TIP & 2013 & SCO & RE & G & SD + C & NL & multiple & NA \\
    \hline  \hline
    \multirow{2}{*}{\#} & \multirow{2}{*}{\textbf{Model}} & \multirow{2}{*}{\textbf{Pub}} & \multirow{2}{*}{\textbf{Year}} & \multirow{2}{*}{\textbf{Cues}} & \multirow{2}{*}{\textbf{Elements}} & \multicolumn{2}{c|}{\textbf{Hypothesis}} & \textbf{Aggregation} & \multirow{2}{*}{\textbf{Source}} & \multirow{2}{*}{\textbf{Code}} \\
    \cline{7-8}
   & & & & & & \textbf{Uniqueness} & \textbf{Prior} & (\textbf{Optimization}) & &  \\
    \hline
    18 & \textbf{LS}~\cite{NiuGLL12stereopsis} & CVPR & 2012 & DP & RE & G & DK & NL & stereo images & NA \\
    19 & \textbf{DRM}~\cite{desingh2013depth} & BMVC & 2013 & DP & RE & G & - & SVM & Kinect & NA \\

    20 & \textbf{SDLF}~\cite{li2014saliency} & CVPR & 2014 & LF & mRE & G & F + B + O & NL & Lytro camera & NA \\
    \hline
  \end{tabular}
  \caption{Salient object detection models with extrinsic cues grouped by their adopted cues. For cues, \{GT = ground-truth annotation, SI = similar images, TC = temporal cues, SCO = saliency co-occurrence, DP = depth, and LF = light field\}. For saliency hypothesis, \{P = generic properties, PRA = pre-attention cues, HD = discriminativity in high-dimensional feature space, SS = saliency similarity,
  CMP = complement of saliency cues, SP = sampling probability,
  MCO = motion coherence, RP = repeatedness, RS = region similarity,
  C = corresponding, and DK = domain knowledge.\}. Others, \{CRF = conditional random field, SVM = support vector machine, BDT = boosted decision tree, and RF = random forest.\}.}
  \label{tab:SalModelExtrinsic}
  \vspace{-10pt}
\end{figure*}


\subsubsection{Models with Extrinsic Cues}\label{sec:groupE}

\newcommand{\calC}{{\mathcal C}}

Models in the third subgroup adopt the \emph{extrinsic cues}
to assist the detection of salient objects in images and videos.
In addition to the visual cues observed from the single input image,
the extrinsic cues can be derived from the ground-truth annotations
of the training images, similar images, the video sequences,
a set of input images containing the common salient objects,
depth maps, or light field images.
In this section, we will review these models according to the
types of used extrinsic cues.
\tabref{tab:SalModelExtrinsic} lists all the models with
extrinsic cues,
where each method is highlighted with several pre-defined attributes.

\textbf{Salient object detection with similar images}.
With the availability of increasingly large amount of
visual content on the web,
salient object detection by leveraging the visually similar
images to the input image has been studied in recent years.
Generally, given the input image $I$, $K$
similar images $\calC_{I} = \{I_k\}_{k=1}^K$ are first retrieved
from a large collection of images $\calC$.
The salient object detection on the input $I$ can be assisted
by examining these similar images.

\newcommand{\calD}{{\mathcal D}}

In some studies, it is assumed that saliency annotations of
$\mathcal{C}$ are available.
For example, Marchesotti~\etal~\cite{marchesotti2009framework}
propose to describe each indexed image $I_k$ by a pair of
descriptors $(\vf_{I_k}^+, \vf_{I_k}^-)$,
where $\vf_{I_k}^+$ and $\vf_{I_k}^-$ denote the feature descriptors
(Fisher vector) of the salient and non-salient regions
according to the saliency annotations, respectively.
To compute the saliency map, each patch $p_x$ of the input image
is described by a fisher vector $\vf_x$.
Saliency of patches are computed according to their contrast with
foreground and background region features
$\{(\vf_{I_k}^+, \vf_{I_k}^-)\}_{k=1}^K$.


Alternatively, based on the observation that different features contribute differently to the saliency analysis on each image, Mai~\etal~\cite{MaiNL13Aggregation} propose to learn the image specific rather than universal weights to fuse the saliency maps that are computed on different feature channels. To this end, the CRF aggregation model of saliency maps is trained only on the retrieved similar images to account for the dependence of aggregation on individual images\footnote{We will discuss more technical details about \cite{MaiNL13Aggregation} in Sect. \ref{sec:groupO}.}.

Saliency based on similar images works well if large-scale image collections are available. Saliency annotation, however, is time consuming, tedious, and even intractable on such collections. To mitigate this, some methods leverage the \emph{unannotated} similar images. With the web-scale image collections $\mathcal{C}$, Wang~\etal~\cite{wang2011image} propose a simple yet effective saliency estimation algorithm. The pixel-wise saliency map is computed as:
\begin{align}
s(x) = \sum\nolimits_{k=1}^K ||I(x) - \tilde{I}_k(x)||_1,
\end{align}
where $\tilde{I}_k$ is the geometrically warped version of $I_k$ with the reference $I$. The main insight is that similar images offer good approximations to the background regions while salient regions might not be well-approximated.

Siva~\etal~\cite{siva2013looking} propose a probabilistic formulation for saliency computation as a sampling problem. A patch $p_x$ is considered to be salient if it has the low probability of being sampled from the images $\calC_I \cup I$. In another word, higher saliency scores will be given to $p_x$ if it is unique among a bag of patches extracted from similar images.

\textbf{Co-saliency object detection}.
Instead of concentrating on computing saliency on a single image,
co-salient object detection algorithms focus on discovering
the \emph{common} salient objects shared by multiple input images
$\{I^i\}_{i=1}^M$.
That is, such objects can be the same object with different
viewpoints or the objects of the same category sharing
similar visual appearances.
Note that the key characteristic of co-salient object
detection algorithms is that their input is
\emph{a set} of images,
while classical salient object detection models only need
a \emph{single} input image.

Co-saliency detection is closely related to the concept of image
co-segmentation that aims to segment similar objects from
multiple images~\cite{rother2006Cosegmentation, batra2010icoseg}.
As stated in \cite{fu2013cluster}, three major differences
exist between co-saliency and co-segmentation.
First, co-saliency detection algorithms only focus on detecting
the common salient objects while the similar
but non-salient background might be also segmented out
in co-segmentation approaches
\cite{mukherjee2011scale, kim2011distributed}.
Second, some co-segmentation methods, \eg~\cite{batra2010icoseg},
need user input to guide the segmentation process in
ambiguous situations.
Third, salient object detection often serves as a
pre-processing step,
and thus more efficient algorithms are preferred than
co-segmentation algorithms, especially over a large number of images.

Li and Ngan~\cite{li2011co} propose a method to compute co-saliency for an image pair with some objects in common. The co-saliency is defined as the inter-image correspondence, \ie~low saliency values should be given to the dissimilar regions.
Similarly in~\cite{ChangLL11from},
Chang~\etal~propose to compute co-saliency by exploiting the
additional \emph{repeatedness} property across multiple images.
Specifically, the co-saliency score of a pixel is defined
as the multiplication of its traditional saliency
score~\cite{goferman2012context} and its repeatedness
likelihood over the input images.
Fu~\etal~\cite{fu2013cluster} propose a cluster-based
co-saliency detection algorithm by exploiting the
well-established global contrast and spatial distribution
concepts on a single image.
Additionally, the corresponding cues over multiple images
are introduced to account for the saliency co-occurrence.


\newcommand{\vp}{{\textbf p}}
\newcommand{\vH}{{\textbf H}}
\newcommand{\calL}{{\mathcal L}}

\subsubsection{Other Classic Models}\label{sec:groupO}
In this section, we review algorithms that aim to directly segment or localize salient objects with bounding boxes, and algorithms that are closely related to saliency detection. Some subsections offer a different categorization of some models covered in the previous sections (e.g., supervised vs. unsupervised). See~\figref{tab:salModelsOthers}.

\textbf{Localization models}.
Liu~\etal~\cite{LiuSZTS07Learn} convert the binary segmentation map to bounding boxes. The final output is a set of rectangles around salient objects. Feng~\etal~\cite{feng2011salient} define saliency for a sliding window as its composition cost using the remaining image parts. Based on an over-segmentation of the image, the local maxima, which can efficiently be found among all sliding windows in a brute-force manner, are assumed to correspond to salient objects.

The basic assumption in many previous approaches is that at least one salient object exists in the input image. This may not always hold as some \emph{background images} contain no salient objects at all. In~\cite{wang2012salient}, Wang~\etal~investigate the problem of localizing and predicting the \emph{existence} of salient objects on thumbnail images. Specifically, each image is described by a set of features extracted in multiple channels. The existence of salient objects is formulated as a binary classification problem. For localization, a regression function is learned using a Random Forest regressor on training samples to directly output the position of the salient object.

\begin{figure}[t]
    \renewcommand{\arraystretch}{.6}
    \renewcommand{\tabcolsep}{2mm}
    \centering
    \small
    \begin{tabular}{|l||l|ll|c|c|} \hline
    \textbf{\#} & \textbf{Model} & \textbf{Pub} & \textbf{Year} & \textbf{Type} &\textbf{Code} \\
    \hline \hline
    1 & \textbf{COMP}~\cite{feng2011salient} & ICCV & 2011 & Localization & NA  \\
    2 & \textbf{GSAL}~\cite{wang2012salient} & CVPR & 2012 & Localization & NA \\
    3 & \textbf{CTXT}~\cite{wang2011automatic} & ICCV & 2011 & Segmentation & NA \\
    4 & \textbf{LCSP}~\cite{tian2014learning} & IJCV & 2014 & Segmentation & NA \\
    5 & \textbf{BENCH}~\cite{borji2012salient} & ECCV & 2012 & Aggregation & M \\
    6 & \textbf{SIO}~\cite{li2013estimating} & SPL & 2013 & Optimization & NA \\
    7 & \textbf{ACT}~\cite{mishra2012active} & PAMI & 2012 & Active & C \\
    8 & \textbf{SCRT}~\cite{liXiaodiCVPR2014} & CVPR & 2014 & Active & NA \\
    9 & \textbf{WISO}~\cite{borjiTIP2014} & TIP & 2014 & Active & NA \\
    \hline
    \end{tabular}
    \caption{Other salient object detection models}.
    \label{tab:salModelsOthers}
    \vspace{-20pt}
\end{figure}


\textbf{Segmentation models}. Segmenting salient objects is closely related to the figure-ground problem, which is essentially a binary classification problem trying to separate the salient object from the background. Yu~\etal\cite{yu2010automatic} utilize the complementary characteristics of imperfect saliency maps generated by different contrast-based saliency models.
Specifically, two complementary saliency maps are first generated for each image, including a sketch-like map and an envelope-like map. The sketch-like map can accurately locate parts of the most salient object (\ie~skeleton with high precision), while the envelope-like map can roughly cover the entire salient object (\ie~envelope with high recall). With these two maps, the reliable foreground and background regions can be detected from each image first to train a pixel classifier. By labeling all other pixels with this classifier, the salient object can be detected as a whole. This method is extended in \cite{tian2014learning} by learning the complementary saliency maps for the purpose of salient object segmentation.

Lu~\etal~\cite{lu2011salient} exploit the convexity (concavity) prior for salient object segmentation. This prior assumes that the region on the convex side of a curved boundary tends to belong to the foreground. Based on this assumption, concave arcs are first found on the contours of superpixels. For a concave arc, its convexity context is defined as the windows which are tightly close to the arc. An undirected weight graph is then built over the superpixels with concave arcs, where the weights between vertices are determined by the summation of concavity context on different scales in the hierarchical segmentation of the image. Finally, the Normalized Cut algorithm~\cite{ShiM00Ncut} is performed to separate the salient object from the background.

To leverage the contextual cues more effectively, Wang~\etal~\cite{wang2011automatic} propose to integrate an auto-context classifier~\cite{TuB10Auto} into an iterative energy minimization framework to automatically segment the salient object. The auto-context model is a multi-layer Boosting classifier on each pixel and its surroundings to predict if it is associated with the target concept. The subsequent layer is built on the classification of the previous layer. Hence through the layered learning process, the spatial context is automatically utilized for more accurate segmentation of the salient object.

\textbf{Supervised vs. unsupervised models}.
The majority of the existing learning-based works on saliency detection focus on the supervised scenario, \ie~learning a salient object detector
given a set of training samples with ground-truth annotations. The aim here is to separate the salient elements from the background elements.

Each element (\eg~a pixel or a region) in the input image is
represented by a feature vector $\textbf{f}\in\mathbb{R}^D$,
where $D$ is the feature dimension.
Such a feature vector is then mapped to a saliency score
$s\in\mathbb{R}^+$ based on the learned linear or non-linear
mapping function $f: \mathbb{R}^D\rightarrow \mathbb{R}^+$.

\newcommand{\vw}{{\textbf w}}
\newcommand{\vx}{{\textbf x}}

One can assume the mapping function $f$ is linear, \ie~$s = \textbf{w}^\text{T}\textbf{f}$, where $\textbf{w}$ denotes the combination weights of all components in the feature vector. Liu~\etal~\cite{LiuSZTS07Learn} propose to learn the weights with the Conditional Random Field (CRF) model trained on the rectangular annotations of the salient objects. In a recent work~\cite{lu2014learning}, the large-margin framework is adopted to learn the weights $\textbf{w}$.

Due to the highly non-linear essence of the saliency mechanism,
however, the linear mapping might not perfectly capture the
characteristics of saliency.
To this end, such a linear integration is extended in
\cite{khuwuthyakorn2010object},
where a mixture of linear Support Vector Machines (SVM) is
adopted to partition the feature space into a set of sub-regions
that are linearly separable using a divide-and-conquer strategy.
In each region, a linear SVM, its mixture weights,
and the combination parameters of the saliency features are
learned for better saliency estimation.
Alternatively, other non-linear classifiers such as boosted decision trees (BDT)~\cite{mehrani2010saliency,kim2014salient} and the random forest (RF)~\cite{JiangWYWZL13} are also utilized.

Generally speaking, supervised approaches allow richer
representations for the elements compared with the heuristic methods.
In the seminal work of the supervised salient object detection,
Liu~\etal~\cite{LiuSZTS07Learn} propose a set of features
including the local multi-scale contrast,
regional center-surround histogram distance,
and global color spatial distribution.
Similar to models with only intrinsic cues,
region-based representation for salient object detection has
become increasingly popular as more sophisticated descriptors
can be extracted at region level.
Mehrani and Veksler~\cite{mehrani2010saliency} demonstrate
promising results by considering generic regional properties,
\eg~color and shape, which are widely used in other applications
like image classification.
Jiang~\etal~\cite{JiangWYWZL13} propose a regional saliency
descriptor including the regional local contrast, regional
backgroundness, and regional generic properties.
In~\cite{lu2014learning, kim2014salient},
each region is described by a set of features such as local
and global contrast, backgroundness, spatial distribution,
and the center prior.
The pre-attentive features are also considered
in~\cite{lu2014learning}.

Usually, the richer representations result in feature vectors with
higher dimensions,
\eg~$D=93$ in~\cite{JiangWYWZL13} and $D=75$
in~\cite{kim2014salient}.
With the availability of a large collections of training samples,
the learned classifier is capable of automatically integrating
such richer features and picking up the most discriminative ones.
Therefore, better performance can be expected compared with
the heuristic methods.


\if 1
\begin{figure}
  \centering
  \footnotesize
\centering
\begin{tabular}{|c||c|cc|c|c|c|}
\hline
\# & Model & Pub  & Year & Local + Global & Fine-tuning & Network \\ \hline
\hline
1  & \textbf{SCNN} \cite{SuperCNN_IJCV2015}  & IJCV & 2015 & Implicit & N & 1D AlexNet\\
2  & \textbf{LEGS} \cite{wang2015Deep}  & CVPR & 2015 & Explicit & N & 3 Conv, 3 FC\\
3  & \textbf{MC} \cite{zhao2015saliency}   & CVPR & 2015 & Explicit & Y & Clarifai\\
4  & \textbf{MDF} \cite{li2015visual}  & CVPR & 2015 & Explicit & N & AlexNet\\
5  & \textbf{DISC} \cite{chen2016DISC} & NNLS & 2016 & Explicit & N & AlexNet\\
6  & \textbf{DCL} \cite{Li2016deep}  & CVPR & 2016 & Implicit & Y & VGGNet\\ \hline
\end{tabular}
\caption{Feature learning based models.}
\label{table:deep_learning}
\vspace{-10pt}
\end{figure}
\fi

Some models have utilized unsupervised techniques. In~\cite{siva2013looking}, saliency computation is formulated in a probabilistic framework as a sampling problem. The saliency of each image patch is proportional to its sampling probability from all of the patches, which are extracted from both the input image and the similar images retrieved from a corpus of unlabeled images.
In~\cite{qin2015saliency}, cellular automata is exploited for unsupervised salient object detection.

\textbf{Aggregation and optimization models}. Given $M$ saliency maps $\{S_i\}_{i=1}^M$, coming from different salient object detection models or hierarchical segmentations of the input image, aggregation models try to form a more accurate saliency map.
Let $S_i(x)$ denote the saliency value of pixel $x$ of the $i$-th saliency map. In~\cite{borji2012salient}, Borji~\etal~propose a standard saliency aggregation method as follows:
\begin{align}
 S(x) = P(s_x = 1|\mathbf{f}_x) \propto \frac{1}{Z} \sum\nolimits_{i=1}^M \zeta(S_i(x))
\end{align}
where $\mathbf{f}_x=(S_1(x), S_2(x), \ldots, S_M(x))$ is the saliency scores for pixel $x$ and $s_x=1$ indicates $x$ is labeled as salient. $\zeta(\cdot)$ is a real-valued function which can take the following form:
\begin{align}
\zeta_1(z) = z; \ \zeta_2(z) = \text{exp}(z); \ \zeta_3(z) = -\frac{1}{\text{log}(z)}.
\end{align}


Inspired by the aggregation model in \cite{borji2012salient}, Mai~\etal~\cite{MaiNL13Aggregation} propose two aggregation solutions. The first solution adopts the pixel-wise aggregation:
\begin{align}
 P(s_x = 1|\mathbf{f}_x; \lambda) = \sigma\left(\sum_{i=1}^M \lambda_i S_i(x) + \lambda_{M+1}\right)
\end{align}
where $\lambda = \{\lambda_i|i = 1,\ldots, M+1\}$ is the set of model parameters and $\sigma(z) = 1/(1+\text{exp}(-z))$. However, it is noted that one potential problem of such direct aggregation is its ignorance of the interaction between neighboring pixels. Inspired by~\cite{liu2011learning}, they propose the second solution by using the CRF to aggregate saliency maps of multiple methods to capture the relation between neighboring pixels. The parameters of the CRF aggregation model are optimized on the training data. The saliency of each pixel is the posterior probability of being labeled as salient with the trained CRF. 

\newcommand{\calS}{{\mathcal S}}

Alternatively, Yan~\etal~\cite{yan2013hierarchical} integrate the saliency maps computed on the hierarchical segmentations of the image into a tree-structure graphical model, where each node corresponds to a region in every hierarchy.
Thanks to the tree structure, the saliency inference can efficiently be conducted using belief propagation. In fact, solving the three layer hierarchical model is equivalent to applying a weighted average to all single-layer maps. Different from naive multi-layer fusion, this hierarchical inference algorithm can select optimal weights for each region instead of global weighting.

Li~\etal~\cite{li2013estimating} propose to optimize the saliency values of all superpixels in an image to simultaneously meet several saliency criteria including visual rarity, center-bias and mutual correlation.
Based on the correlations (similarity scores) between region pairs,
the saliency value of each superpixel is optimized by quadratic programming
when considering the influences of all the other superpixels.
Let $w_{ij}$ denote the correlation between two regions $r_i$ and $r_j$,
the saliency values $\{s_i\}_{i=1}^N$ (denoted by $s(r_i)$ as $s_i$ for short) can be optimized by solving:
\begin{equation}\label{Eq:objective}
\begin{split}
\min_{\{s_i\}_{i=1}^N}&\sum_{i=1}^{N}s_{i}\sum_{j\neq{}i}^{N}w_{ij}+\lambda_c\sum_{i=1}^{N}s_{i}e^{{d_i}/{d_D}}\\
&+\lambda_r\sum_{i=1}^{N}\sum_{j\neq{}i}^{N}(s_i-s_j)^2w_{ij}e^{-{d_{ij}}/{d_D}}\\
s.t.~ ~&0\leq{}s_i\leq1,\forall i,~~\text{and}~\sum_{i=1}^{N}{s_i}=1.
\end{split}
\end{equation}
where $d_D$ is half the image diagonal length. $d_{ij}$ and $d_i$ are the spatial distances from the $r_i$ to $r_j$ and the image center, respectively. In the optimization, the saliency value of each superpixel is optimized by quadratic programming when considering the influences of all other superpixels.
%
Zhu~\etal\cite{zhu2014saliency} also adopt a similar optimization-based framework to integrate multiple foreground/background cues as well as the smoothness terms to automatically infer the optimal saliency values.

The Bayesian framework is adopted to more effectively integrate the complementary dense and sparse reconstruction errors~\cite{li2013saliency}. A fully-connected Gaussian Markov Random Field between each pair of regions is constructed to enforce the consistency between salient regions~\cite{jia2013category}, which leads to an efficient computation of the final regional saliency scores.

\textbf{Active models.}
Inspired by the interactive segmentation models (\eg~\cite{LiSTS04Lazy,RotherKB04Grab}), a new trend has emerged recently by explicitly decoupling the two stages of saliency detection mentioned in Sec.~\ref{whatIS}: a) detecting the most salient object and b) segmenting it. Some studies propose to perform active segmentation by utilizing the advantages of both fixation prediction and segmentation models. For example, Mishra~\etal~\cite{mishra2012active} combine multiple cues (\eg~color, intensity, texture, stereo and/or motion) to predict fixations. The ``optimal'' closed contour for salient object around the fixation point is then segmented in polar space. Li~\etal~\cite{liXiaodiCVPR2014} propose a model composed of two components: a \textit{segmenter} that
proposes candidate regions and a \textit{selector} that gives each region a saliency score (using a fixation prediction model). Similarly, Borji~\cite{borjiTIP2014} proposes to first roughly locate the salient object at the peak of the fixation map (or its estimation using a fixation prediction model) and then segment the object using superpixels. The last two algorithms adopt annotations to determine the upper-bound of the segmentation performance, propose datasets with multiple objects in scenes, and provide new insight to the inherent connections of fixation prediction and salient object segmentation.

\textbf{Salient object detection on videos}.
In addition to the spatial information,
video sequence provides the temporal cue, \eg~
motion which facilitates salient object detection.
Zhai and Shah~\cite{zhai2006visual} first estimate the
keypoint correspondences between two consecutive frames.
Motion contrast is computed based on the planar motions
(homography) between images,
which is estimated by applying RANSAC on point correspondences.
%
%
Liu~\etal~\cite{liu2008video} extend their spatial saliency features
\cite{LiuSZTS07Learn} to the motion field resulting from the optical
flow algorithm.
With the colorized motion field as the input image,
the local multi-scale contrast, regional center-surround distance,
and global spatial distribution are computed and finally integrated
in a linear way.
Rahtu~\etal~\cite{rahtu2010segmenting} integrate the spatial saliency
into the energy minimization framework by considering
the temporal coherence constraint.
Li~\etal~\cite{bin2013temporally} extend the regional contrast-based
saliency to the spatio-temporal domain.
Given the over-segmentation of the frames of the video sequence,
spatial and temporal region matchings between each two consecutive
frames are estimated based on their color, texture,
and motion features in a interactive manner on an undirected
un-weighted matching graph.
The saliency of a region is determined by computing its local contrast
to the surrounding regions not only in the present frame
but also in the temporal domain.

\textbf{Salient object detection with depth}.
We live in real 3D environments
where stereoscopic content provide additional depth cues for guiding visual attention and understanding the surroundings. 
This is further validated by Lang~\etal~\cite{LangNKYKY12Depth} through
experimental analysis of the importance of depth cues for
eye fixation prediction.
Recently, researchers have started to study how to exploit the
depth cues for salient object detection
\cite{NiuGLL12stereopsis, desingh2013depth},
which might be captured indirectly from the stereo images or directly
using a depth camera (\eg~Kinect).

The most straightforward extension is to adopt the widely used
hypotheses introduced in \secref{sec:groupEI} and~\ref{sec:groupLI}
to the depth channel, \eg~the global contrast on the depth map~\cite{NiuGLL12stereopsis, desingh2013depth}. Furthermore, Niu~\etal~\cite{NiuGLL12stereopsis} demonstrate how to leverage the domain knowledge in stereoscopic photography to compute the saliency map. The input image is first segmented into regions $\{r_i\}$. In practice, the attended regions are often assigned small or zero disparities to minimize the \emph{vergence-accommodation conflict}. Therefore, the first type of regional saliency based on the disparity is defined as:
\begin{align}
s_{d, 1}(r_i) = \left\{\begin{array}{cc}
            \frac{d_{max}-\bar{d_i}}{d_{max}} & \mbox{if $\bar{d_i}\geq 0$} \\
            \frac{d_{min}-\bar{d_i}}{d_{min}} & \mbox{if $\bar{d_i}< 0$}
          \end{array}
        \right.
\end{align}
where $d_{max}$ and $d_{min}$ are the maximal and minimal disparities, respectively. $\bar{d_i}$ denotes the average disparity in region $r_i$.
Additionally, objects with negative disparities are perceived popping out from the scene. The second type of regional stereo saliency is then defined as:
\begin{align}
s_{d, 2}(r_i) = \frac{d_{max} - \bar{d_i}}{d_{max} - d_{min}}.
\end{align}
Stereo saliency is linearly computed by an adaptive weight.


\textbf{Salient object detection on light field}. The idea of using light field for saliency detection was proposed in~\cite{li2014saliency}. A light field, captured using a specifically designed camera \eg~Lytro, is essentially an array of images shot by a grid of cameras viewing the scene. The light field data offers two benefits for salient object detection: 1) it allows synthesizing a stack of images focusing at different depths, and 2) it provides an approximation of scene depth and occlusions.

With this additional information, Li~\etal~\cite{li2014saliency} first utilize the focusness and objectness priors to robustly choose the background and select the foreground candidates. Specifically, the layer with the estimated background likelihood score is used to estimate the background regions. Regions, coming from Mean-shift algorithm, with the high foreground likelihood score are chosen as salient object candidates. Finally, the estimated background and foreground are utilized to compute the contrast-based saliency map on the all-focus image.

A new challenging benchmark dataset for light-field saliency analysis, known as HFUT-Lytro, has been recently introduced in~\cite{zhang2017saliency}. 

\subsection{New Testament: Deep Learning Based Models}

All the methods that we have reviewed so far aim at detecting salient objects using heuristics.
While hand-crafted features allow real-time detection performance, they suffer from several shortcomings that limit their ability in capturing salient objects in challenging scenarios.

Convolutional neural networks (CNNs) \cite{lecun1998gradient},
as one of the most popular tools in machine learning, have been
applied to many vision problems such as object recognition~\cite{krizhevsky2012imagenet}, semantic segmentation~\cite{long2015fully} and edge detection\cite{xie2015holistically}.
%
%
Recently, it has been shown that CNNs \cite{li2015visual,
SuperCNN_IJCV2015} are also very effective when applied to salient
object detection.
Thanks to their multi-level and multi-scale features, CNNs are
capable of accurately capturing the most salient regions
without using any prior knowledge (\eg~segment-level information).
Furthermore, multi-level features allow CNNs to better
locate the boundaries of the detected salient regions, even when  shades or reflections exist.
By exploiting the strong feature learning ability of CNNs, a series of algorithms are
proposed to learn the saliency representations from large amounts
of data.
These CNN-based models continuously refresh the records on
almost all existing datasets and are becoming the main stream solution.
The rest of this subsection is dedicated to reviewing CNN-based models.

Basically, salient object detection models based on deep learning can be split into two main categories.
The first category includes models that have used multi-layer perceptrons (MLPs) for saliency detection.
In these models, the input image is usually over-segmented into single- or multi-scale small regions.
Then, a CNN is used to extract high-level features which are later fed to a MLP to determine the saliency value of a small region.
Though high-level features are extracted from CNNs, unlike fully convolutional networks (FCNs), the spatial information from CNN features cannot be preserved because of the utilization of MLPs.
To highlight the differences between these methods and FCN-based methods, we call them "Classic Convolutional Network based" (CCN-based) methods.
%
%
The second category includes models that are based on
"Fully Convolutional Networks" (FCN-based).
The pioneering work of Long \etal \cite{long2015fully} falls under this category and
aims at solving the semantic segmentation problem.
Since salient object detection is inherently a segmentation task, a number of researchers have adopted FCN-based architectures because of their capability in preserving spatial information.

\figref{tab:salientObjModelsCNNs} shows a list of CNN-based saliency models.

\begin{figure*}[thp!]
  \centering
  \small
  \renewcommand{\arraystretch}{.8}
  \renewcommand{\tabcolsep}{2.5mm}
  \begin{tabular}{|c||l|cc|c|c|c|c|c|}
  \hline
  \multirow{2}{*}{\#} & \multirow{2}{*}{\textbf{Model}} & \multirow{2}{*}{\textbf{Pub}} & \multirow{2}{*}{\textbf{Year}} & \multirow{2}{*}{\textbf{\#Training Images}} & \multirow{2}{*}{\textbf{Training Set}} & \multirow{2}{*}{\textbf{Pre-trained Model}} & \multirow{2}{*}{\textbf{Fully Conv}}  \\
   & & & & & & & \\
  \hline \hline
  1 & \textbf{SuperCNN}~\cite{SuperCNN_IJCV2015} & IJCV & 2015 & 800 & ECSSD & - & \xmark \\
  2 & \textbf{LEGS}~\cite{wang2015Deep} & CVPR & 2015 & 3,340 & MSRA-B + PASCALS & - & \xmark \\
  3 & \textbf{MC}~\cite{zhao2015saliency} & CVPR & 2015 & 8,000 & MSRA10K & GoogLeNet \cite{szegedy2015going} & \xmark \\
  4 & \textbf{MDF}~\cite{li2015visual} & CVPR & 2015 & 2,500 & MSRA-B & - & \xmark \\
  5 & \textbf{HARF}~\cite{zou2015harf} & ICCV & 2015 & 2,500 & MSRA-B & - & \xmark \\


  6 & \textbf{ELD}~\cite{lee2016deep} & CVPR & 2016 & nearly 9,000 & MSRA10K & VGGNet & \xmark \\

  7 & \textbf{SSD-HS}~\cite{kim2016ECCV} & ECCV & 2016 & 2,500 & MSRA-B & AlexNet & \xmark \\
  8 & \textbf{FRLC}~\cite{wang2016salient} & ICIP & 2016 & 4,000 & DUT-OMRON & VGGNet & \xmark \\

  9 & \textbf{SCSD-HS}~\cite{kim2016ICPR} & ICPR & 2016 & 2,500 & MSRA-B & AlexNet & \xmark \\
  10 & \textbf{DISC}~\cite{chen2016disc} & TNNLS & 2016 & 9,000 & MSRA10K & - & \xmark \\

  11 & \textbf{LCNN}~\cite{li2017cnn} & Neuro & 2017 & 2,900& MSRA-B + PASCALS & AlexNet & \xmark \\ \hline \hline

  12 & \textbf{DHSNET}~\cite{liu2016dhsnet} & CVPR & 2016 & 6,000 & MSRA10K & VGGNet & \cmark \\
  13 & \textbf{DCL}~\cite{li2016deepX} & CVPR & 2016 & 2,500 & MSRA-B & VGGNet \cite{simonyan2014very} & \cmark \\
  14 & \textbf{RACDNN}~\cite{kuen2016recurrent} & CVPR & 2016 & 10,565 & DUT+NJU2000+RGBD& VGG & \cmark \\
  15 & \textbf{SU}~\cite{kruthiventi2016saliency} & CVPR & 2016 & 10,000 & MSRA10K & VGGNet & \cmark \\
  16 & \textbf{CRPSD}~\cite{tang2016saliency} & ECCV & 2016 & 10,000 & MSRA10K & VGGNet & \cmark \\
  17 & \textbf{DSRCNN}~\cite{tang2016deeply} & MM & 2016 & 10,000 & MSRA10K & VGGNet & \cmark \\
  18 & \textbf{DS}~\cite{li2016deepsaliency} & TIP & 2016 & nearly 10,000 & MSRA10K & VGGNet & \cmark \\
  19 & \textbf{IMC}~\cite{zhang2017deep} & WACV & 2017 & nearly 6,000 & MSRA10K & ResNet & \cmark \\
  20 & \textbf{MSRNet}~\cite{li2017instance} & CVPR & 2017 & 2,500 & MSRA-B + HKU-IS & VGGNet & \cmark \\
  21 & \textbf{DSS}~\cite{hou2016deeply} & CVPR & 2017 & 2,500 & MSRA-B & VGGNet & \cmark \\
  \hline
  \end{tabular}
  \caption{CNN-based salient object detection models and their used information during training. Models in the top part are all CCN-based while models in the bottom part are all FCN-based.
  }\label{tab:salientObjModelsCNNs}
 \vspace{-15pt}
\end{figure*}

\subsubsection{CCN-based Models} \label{sec:mpl_models}

\textbf{One-dimensional (1D) convolution based methods.} As an early attempt, He \etal \cite{SuperCNN_IJCV2015} followed a region-based approach to learn superpixel-wise feature representations.
Their approach dramatically reduces the computational cost compared to pixel-wise CNNs, meanwhile takes global context into consideration.
However, representing a superpixel with the mean color is not informative enough. Further, the spatial structure of the image is difficult to be fully recovered using 1D convolution and pooling operations, leading to cluttered predictions, especially when the input image is a complex scene.

\textbf{Leveraging local and global context.}
Wang \etal~consider both local and global information for better detection of salient regions~\cite{wang2015deepX}.
To this end, two subnetworks are designed for local estimation and global search, respectively.
A deep neural network (DNN-L) is first used to learn local patch features to determine the saliency value of each pixel, followed by a refinement operation which captures the high-level objectness.
For global search, they train another deep neural network (DNN-G) to predict the saliency value of each salient region using a variety of global contrast features such as geometric information, global contrast features, etc.
The top $K$ candidate regions are utilized to compute the final saliency map using a weighted summation.

In \cite{zhao2015saliency}, similar to most of the classic salient object detection methods, both local context and global context are taken into account for constructing a multi-context deep learning framework.
The input image is first fed to the global-context branch to extract global contrast information.
Meanwhile, each image patch which is a superpixel-centered window, is fed to the local-context branch for capturing local information.
A binary classifier is finally used to determine the saliency value by minimizing a unified softmax loss between the prediction value and the ground truth label.
A task-specific pre-training scheme is adopted to jointly optimize the designed multi-context model.

Lee \etal \cite{lee2016deep} exploit two subnetworks to encode low-level and high-level features separately.
They first extract a number of features for each superpixel and feed them into a subnetwork composed of a stack of convolutional layers with $1\times1$ kernel size.
Then, the standard VGGNet \cite{simonyan2014very} is used to capture high-level features.
Both low- and high-level features are flattened, concatenated, and finally fed into a two-layer MLP to judge the saliency of each query region.

\textbf{Bounding box based methods.}
In \cite{zou2015harf}, Zou \etal~proposes a hierarchy-associated rich feature (HARF) extractor.
To do so, a binary segmentation tree is first built for extracting hierarchical image regions and for analyzing the relationships between all pairs of regions. Two different methods are then used to compute two kinds of features ($\text{HARF}_1$ and $\text{HARF}_2$) for regions at the leaf-nodes of the binary segmentation tree.
They leverage all the intermediate features extracted from RCNN \cite{girshick2014rich}, to capture various characteristics of each image region.
With these high-dimensional elementary features, both local regional contrasts and border regional contrasts for each elementary feature type are computed for building a more compact representation.
Finally, the AdaBoost algorithm is adopted to gradually assemble weak decision trees to construct a composite strong regressor.

Kim \etal \cite{kim2016ECCV} design a two-branch CNN architecture to obtain the coarse- and fine representations of the coarse-level and fine-level patches, respectively.
The selective search \cite{uijlings2013selective} method is utilized to generate a number of region candidates that are treated as the input to the two-branch CNN.
Feeding the concatenation of the feature representations of the two branches into the final fully connected layer allows a coarse continuous map to be predicted.
To further refine the coarse prediction map, a hierarchical segmentation method is used to sharpen the boundaries and improve the spatial consistency.

In \cite{wang2016salient}, Wang \etal~solve the salient object detection by employing the Fast R-CNN \cite{girshick2014rich} framework.
The input image is first segmented into multi-scale regions using both over-segmentation and edge-preserving methods.
For each region, the external bounding box is used and the enclosed region is fed to the Fast R-CNN.
A small network composed of multiple fully connected layers is connected to the ROI pooling layer for determining the saliency value of each region.
Finally, an edge-based propagation method is proposed to suppress the background regions and make the resulting saliency map more uniform.

Kim \etal \cite{kim2016ICPR} train a CNN to predict the saliency shape of each image patch.
The selective search method is first used to localize a stack of images patches, each of which is taken as the input to the CNN.
After predicting the shape of each patch, an intermediate mask $M_I$ is computed by accumulating the product of the mask of the predicted shape class and the corresponding probability and averaging all the region proposals.
To further refine the coarse prediction map, a shape class-based saliency detection with hierarchical segmentation (SCSD-HS) is used to incorporate more global information, which is often needed for saliency detection.


\begin{figure*}[tp!]
    \centering  
    \includegraphics[width=0.8\linewidth]{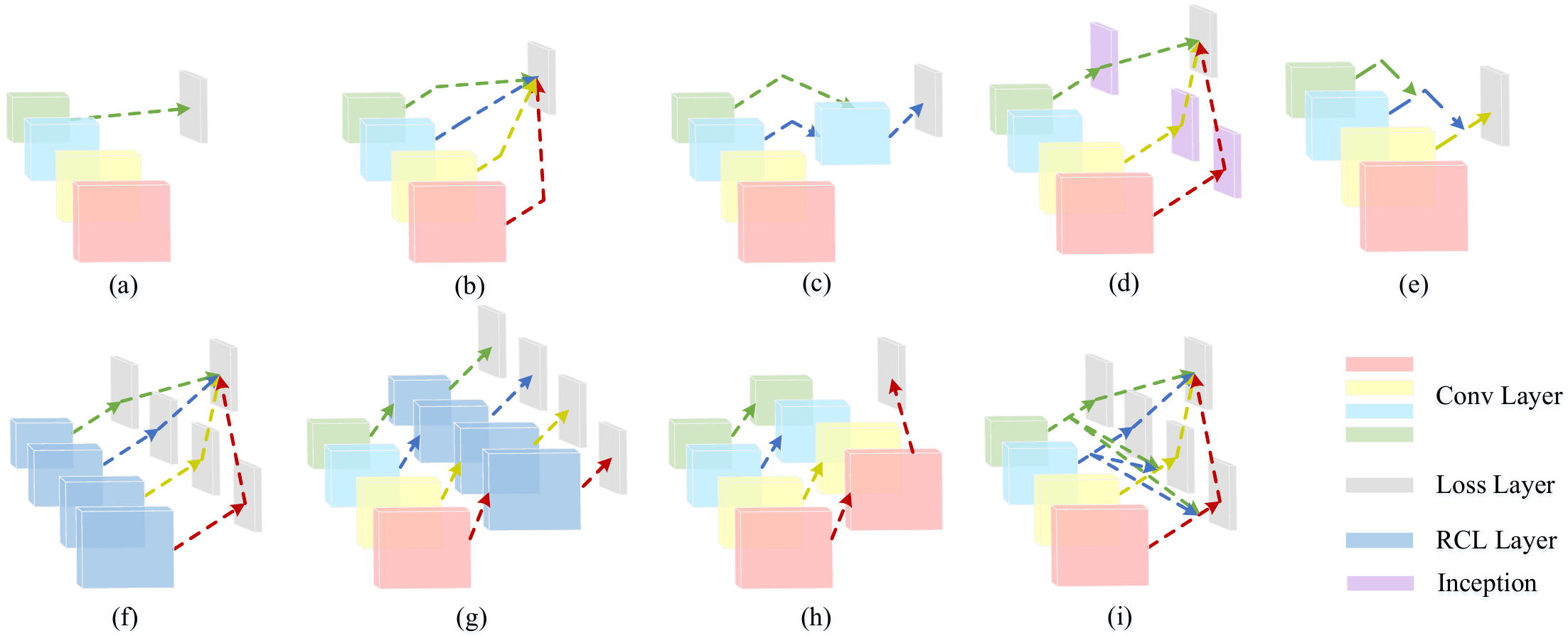} \\

    \caption{Popular FCN-based architectures. One can see that apart from the classical architecture (a) more and more advanced architectures have been developed recently. Some of them (b,c,d,and e) exploit skip layers from different scales so as to learn multi-scale and multi-level features. Some of them (e, g, h, and i) adopt the encoder-decoder structure to better fuse high-level features with low-level ones. There are also some works (f, g, and i) introduce side supervision as done in \cite{xie2015holistically} in order to capture more detailed multi-level information. See Table~\ref{tab:fcn_info} for details on these architectures.
    }\label{fig:fcn_arch}
    \vspace{-10pt}
\end{figure*}

Li \etal \cite{li2017cnn} leverage both high-level features from CNNs and low-level features extracted based on hand-crafted methods.
To enhance the generalization and learning ability of CNNs, the original R-CNN is redesigned by adding local response normalization (LRN) to the first two layers.
%
The selective search method is utilized \cite{uijlings2013selective} to generate a stack of squared patches as the input to the network.
Both high-level and low-level features are fed to a SVM with the $L_1$ hinge-loss to help judge the saliency of each squared region.

\textbf{Models with multi-scale inputs.}
Li \etal \cite{li2015visual} utilize a pre-trained CNN as a feature extractor.
Given an input image, they first decompose it into a series of non-overlapping regions and then feed them into a CNN with three different-scale inputs.
Three subnetworks are then employed to capture advanced features at different scales.
The features obtained from patches at three scales are concatenated and then fed into a small MLP with only two fully connected layers as a regressor to output a distribution over binary saliency labels.
To solve the problem of imperfect over-segmentation, a superpixel based saliency refinement method is used.

\figref{fig:fcn_arch} illustrates a number of popular FCN-based architectures.
\figref{tab:fcn_info} lists different types of information leveraged by these architectures.

\begin{figure}[t]
  \centering
  \small
  \renewcommand{\arraystretch}{.8}
  \renewcommand{\tabcolsep}{1.5mm}
  \begin{tabular}{|c||l|c|c|c|c|c|c|c|}
  \hline
  \multirow{2}{*}{\#} & \multirow{2}{*}{\textbf{Model}} & \multirow{2}{*}{\textbf{SP}} & \multirow{2}{*}{\textbf{SS}} & \multirow{2}{*}{\textbf{RCL}} & \multirow{2}{*}{\textbf{PCF}} & \multirow{2}{*}{\textbf{IL}} & \multirow{2}{*}{\textbf{CRF}} & \multirow{2}{*}{\textbf{Arch.}} \\
   & & & & & & & & \\
  \hline \hline
  1 & \textbf{DCL}~\cite{li2016deepX} & \cmark & \cmark & \xmark & \cmark & \xmark & \cmark & \figref{fig:fcn_arch}(b)\\
  2 & \textbf{CRPSD}~\cite{tang2016saliency} & \cmark & \xmark & \xmark & \xmark & \xmark & \xmark & \figref{fig:fcn_arch}(c) \\
  3 & \textbf{DSRCNN}~\cite{tang2016deeply} & \xmark & \cmark & \cmark & \cmark & \xmark & \xmark & \figref{fig:fcn_arch}(f) \\
  4 & \textbf{DHSNET}~\cite{liu2016dhsnet} & \xmark & \cmark & \cmark & \cmark & \xmark & \xmark & \figref{fig:fcn_arch}(g) \\
  5 & \textbf{RACDNN}~\cite{kuen2016recurrent} & \xmark & \xmark & \cmark & \cmark & \xmark & \xmark & \figref{fig:fcn_arch}(h) \\
  6 & \textbf{SU}~\cite{kruthiventi2016saliency} & \xmark & \cmark & \xmark & \cmark & \xmark & \cmark & \figref{fig:fcn_arch}(d) \\
  7 & \textbf{DS}~\cite{li2016deepsaliency} & \cmark & \xmark & \xmark & \xmark & \xmark & \xmark & \figref{fig:fcn_arch}(a) \\
  8 & \textbf{IMC}~\cite{zhang2017deep} & \cmark & \xmark & \xmark & \xmark & \xmark & \xmark & \figref{fig:fcn_arch}(a) \\
  9 & \textbf{MSRNet}~\cite{li2017instance} & \cmark & \xmark & \xmark & \cmark & \cmark & \cmark & \figref{fig:fcn_arch}(h) \\
  10 & \textbf{DSS}~\cite{hou2016deeply} & \xmark & \cmark & \xmark & \cmark & \xmark & \cmark & \figref{fig:fcn_arch}(i) \\
  \hline
  \end{tabular}
  \caption{Different types of information leveraged by existing FCN-based models. Acronyms include SP: Superpixel, SS: Side Supervision, RCL: Recurrent Convolutional Layer, PCF: Pure CNN Feature, IL: Instance-Level, Arch: Architecture.
  }\label{tab:fcn_info}
  \vspace{-10pt}
\end{figure}

\textbf{Discussion.} As can be seen, the above mentioned MLP-based works rely mostly on segment-level information (e.g., image patches) and classification networks.
These image patches are normally resized to a fixed size and are then fed to a classification network which is used to determine the saliency of each patch.
Some of the models use multi-scale inputs to extract features
in several scales.
However, such a learning framework cannot fully leverage high-level semantic information.
%
Further, spatial information cannot be propagated to the last fully connected layers, thus resulting in global information loss.
%

\subsubsection{FCN-based Models}

Unlike CCN-based models that operate at the patch level, fully convolutional networks (FCNs) \cite{long2015fully} consider pixel-level operations to overcome the problems caused by fully connected layers such as blurriness and inaccurate predictions near the boundaries of salient objects.
Due to desirable properties of FCNs, a great number of FCN-based salient object detection models have been introduced recently.

Li \etal \cite{li2016deepX} design a CNN with two complementary branches: \textit{a pixel-level fully convolutional stream (FCS) and a segment-wise spatial pooling stream (SPS)}.
The FCS introduces a series of skip layers after the last convolutional layer of each stage and then the skip layers are fused together as the output of FCS. Notice that a stage of a CNN is composed of all the layers sharing the same resolution.
The SPS leverages segment-level information for spatial pooling.
Finally, the outputs of FCS and SPS are fused together, followed by a balanced sigmoid cross entropy loss layer as done in \cite{xie2015holistically}.

Liu \cite{liu2016dhsnet} propose two subnetworks to produce a prediction map in a coarse-to-fine and global-to-local manner.
The first subnetwork can be considered as an encoder whose goal is to generate a coarse global prediction.
Then, a refinement subnetwork composed of a series of recurrent convolution layers is used to refine the coarse prediction map from coarse scales to fine scales.

In \cite{tang2016saliency}, Tang \etal~consider both region-level saliency estimation and pixel-level saliency prediction.
For pixel-level prediction, two side paths are connected to the last two stages of the VGGNet and then concatenated for learning multi-scale features.
For region-level estimation, each given image is first over-segmented into multiple superpixels and then the Clarifai model \cite{zeiler2014visualizing} is used to predict the saliency of each superpixel.
The original image and the two prediction maps are taken as the inputs to a small CNN to generate a more convincing saliency map as the final output.

Tang \etal \cite{tang2016deeply} take the deeply supervised net \cite{lee2015deeply} and adopt a similar architecture as in the holistically-nested edge detector \cite{xie2015holistically}.
Unlike HED, they replace the original convolutional layers in VGGNet with recurrent convolutional layers to learn local, global, and contextual information.

In \cite{kuen2016recurrent}, Kuen \etal~propose a two-stage CNN by utilizing spatial transformer and recurrent network units.
A convolutional-deconvolutional network is first used to produce an initial coarse saliency map.
The spatial transformer network \cite{jaderberg2015spatial} is applied to extract multiple sub-regions from the original images, followed by a series of recurrent network units to progressively refine the predictions of these sub-regions.

Kruthiventi \etal~ \cite{kruthiventi2016saliency} consider both fixation prediction and salient object detection in a unified network.
To capture multi-scale semantic information, four inception modules \cite{szegedy2015going} are introduced which are connected to the output of the 2nd, 4th, 5th, and 6th stages, respectively.
These four side paths are concatenated together and passed through a small network composed of two convolutional layers for reducing the aliasing effect of upsampling.
Finally, the sigmoid cross entropy loss is used to optimize the model.

Li \etal \cite{li2016deepsaliency} consider joint semantic segmentation and salient object detection.
Similar to the FCN work \cite{long2015fully}, the two original fully connected layers in VGGNet \cite{simonyan2014very} are replaced by convolutional layers.
To overcome the fuzzy object boundaries caused by the down-sampling operations of CNNs, they make use of the SLIC \cite{achanta2012slic} superpixels to model the topological relationships among superpixels in both spatial and feature dimensions.
Finally, the graph Laplacian regularized nonlinear regression is used to change the combination of the predictions from CNNs and the superpixel graph from the coarse level to the fine level.

Zhang \etal \cite{zhang2017deep} detect salient objects using saliency cues extracted by CNNs and a multi-level fusion mechanism.
The Deeplab \cite{chen2016deeplab} architecture is first used to capture high-level features.
To address the problem of large strides in Deeplab, a multi-scale binary pixel labeling method is adopted to improve spatial coherence, similar to \cite{li2015visual}.

The MSRNet \cite{li2017instance} by Li \etal~consider both salient object detection and instance-level salient object segmentation.
A multi-scale CNN is used to simultaneously detect salient regions and contours.
For each scale, features from upper layers are merged with features from lower layers to gradually refine the results.
To generate a contour map, the MCG \cite{arbelaez2014multiscale} approach is used to extract a small number of candidate bounding boxes and well-segmented regions that are used to help generate salient object instance segmentation.
Finally, a fully connected CRF model \cite{krahenbuhl2011efficient} is employed for refining the spatial coherence.

Hou \etal \cite{hou2016deeply} design a top-down model based on the HED architecture \cite{xie2015holistically}.
Unlike connecting independent side paths to the last convolutional layer of each stage, a series of short connections are introduced to build a strong relationship between each pair of side paths.
As a result, features from upper layers with strongly semantic information are propagated to lower layers, helping them accurately locate exact positions of salient objects.
In the meantime, rich detailed information from lower layers allow the irregular prediction maps from deeper layers to be refined.
A special fusion mechanism is exploited to better combine the saliency maps predicted by different side paths.

\textbf{Discussion}. The foregoing approaches are all based on fully convolutional networks, which enable the point-to-point learning and end-to-end training strategies.
Compared with CCN-based models, these methods make better use of the convolution operation and substantially decrease the time cost.
More importantly, recent FCN-based approaches \cite{hou2016deeply,li2017instance} that utilize CNN features greatly outperform those methods with segment-level information.
%
%

To sum up, the 3 following advantages have been obtained in utilizing FCN-based models for saliency detection.

\noindent  \textbf{1) Local vs. global}. As was mentioned in \secref{sec:mpl_models}, earlier CNN-based models incorporate both local and global contextual information explicitly (embedded in separate networks \cite{wang2015Deep,zhao2015saliency,li2015visual}) or implicitly (using an end-to-end framework). This indeed agrees with the design principles behind many hand-crafted cues reviewed in previous sections. However, FCN-based methods are capable of learning both local and global information internally. Lower layers tend to encode more detailed information such as edge and fine components, while deeper layers favor global and semantically meaningful information. Such properties enable FCN-based networks to drastically outperform classic methods.

\noindent  \textbf{2) Pre-training and fine-tuning}. The effectiveness of fine-tuning a pre-trained network has been demonstrated in many different applications. The network is typically pre-trained on the ImageNet dataset \cite{russakovsky2015imagenet} for image classification. The learned knowledge can be applied to several different target tasks (e.g., object detection \cite{girshick2014rich}, object localization \cite{oquab2014learning}) through simple fine-tuning. A similar strategy has been adopted in salient object detection \cite{zhao2015saliency,li2016deepX} and has resulted in superior performance compared to training from scratch. The learned features, more importantly, are able to capture high-level semantic knowledge on object categories, as the employed networks are pre-trained for scene and object classification tasks.

\noindent  \textbf{3) Versatile architectures}. A CNN architecture is formed by a stack of distinct layers that transform the input images into an output map through a differentiable function. The diversity of FCNs allows designers to design different structures that are appropriate for them.

Despite a great success, FCN-based models still fail in several cases.
Typical examples include scenes with transparent objects, low contrast between foreground and background, and complex backgrounds, as shown in \cite{hou2016deeply}.
This calls for developing of more powerful architectures in the future.

\begin{figure}[tp!]
  \centering
  \footnotesize
    \begin{tabular}{cccccc} 
    \includegraphics[width=0.155\linewidth]{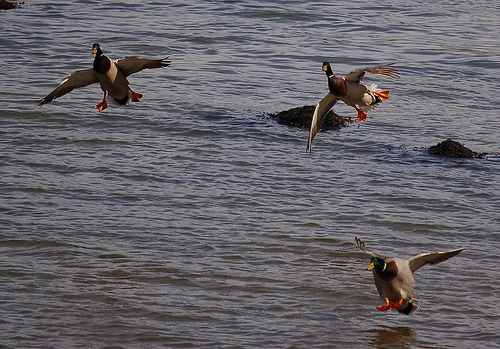}  &
        \includegraphics[width=0.155\linewidth]{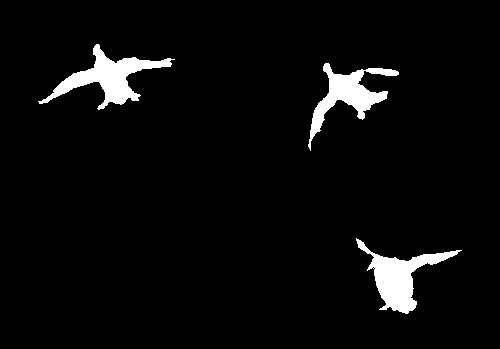} &
        \includegraphics[width=0.155\linewidth]{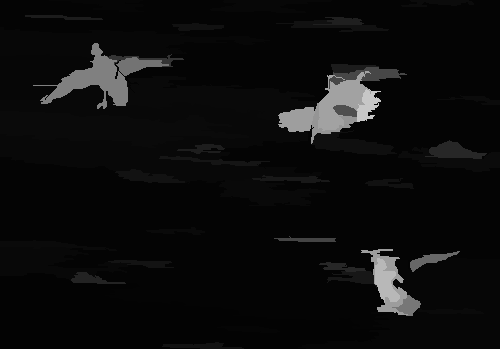} &
        \includegraphics[width=0.155\linewidth]{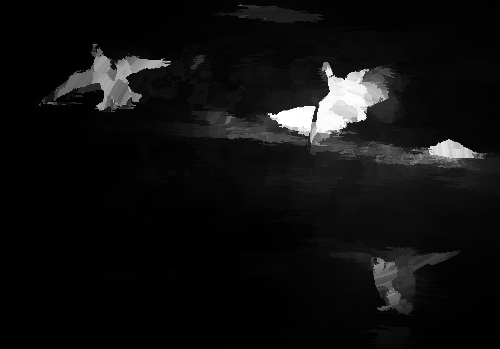} &
        \includegraphics[width=0.155\linewidth]{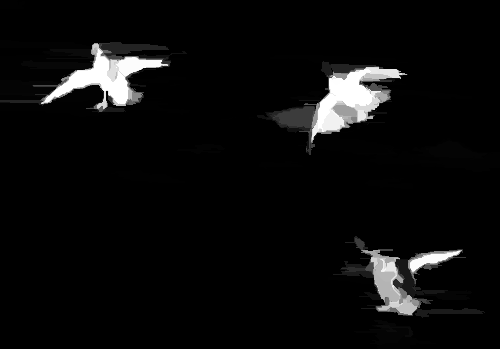} &
        \includegraphics[width=0.155\linewidth]{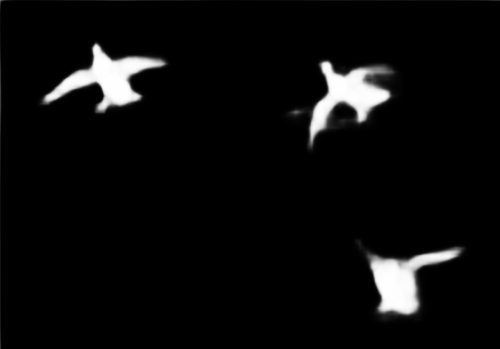} \\
        \includegraphics[width=0.155\linewidth]{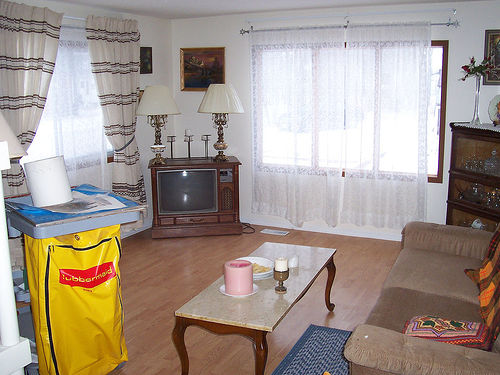}  &
        \includegraphics[width=0.155\linewidth]{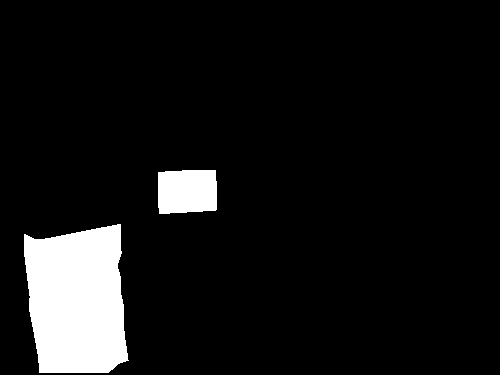} &
        \includegraphics[width=0.155\linewidth]{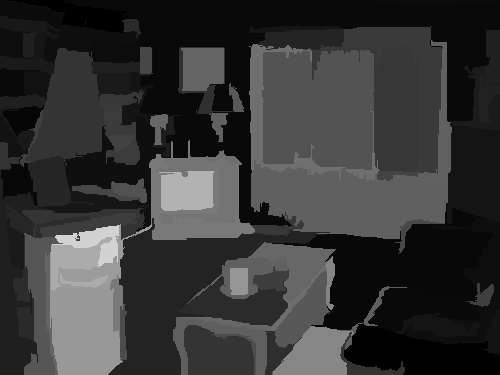} &
        \includegraphics[width=0.155\linewidth]{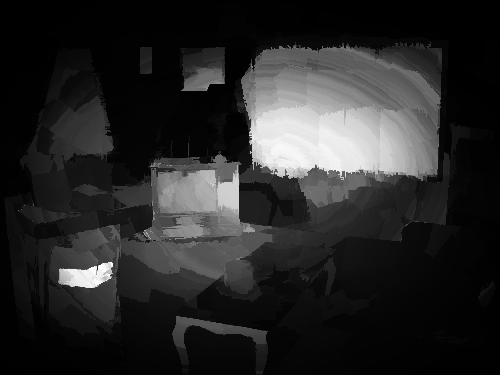} &
        \includegraphics[width=0.155\linewidth]{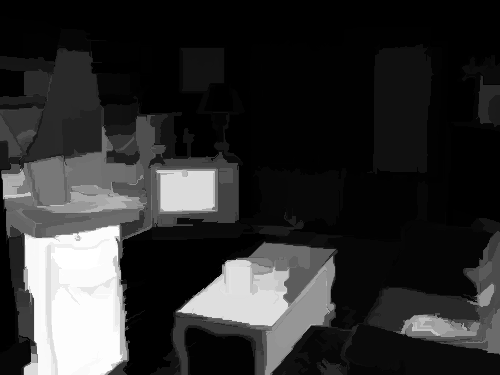} &
        \includegraphics[width=0.155\linewidth]{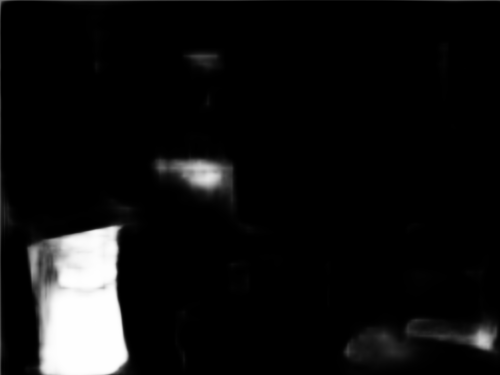} \\
        \includegraphics[width=0.155\linewidth]{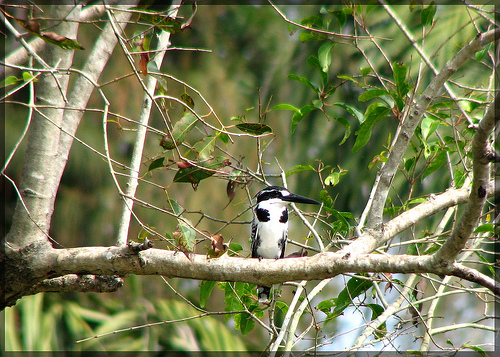}  &
        \includegraphics[width=0.155\linewidth]{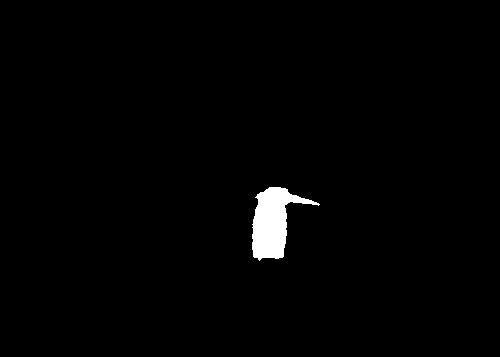} &
        \includegraphics[width=0.155\linewidth]{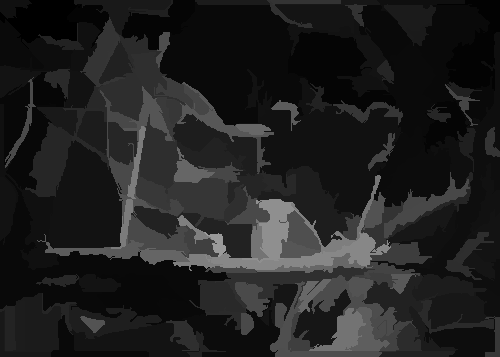} &
        \includegraphics[width=0.155\linewidth]{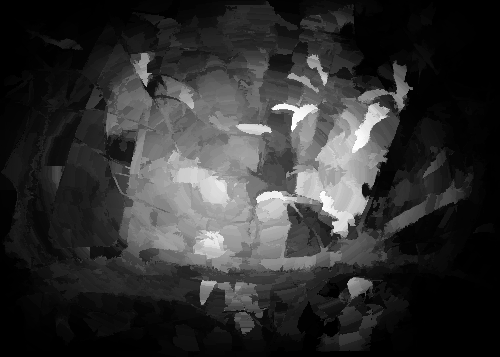} &
        \includegraphics[width=0.155\linewidth]{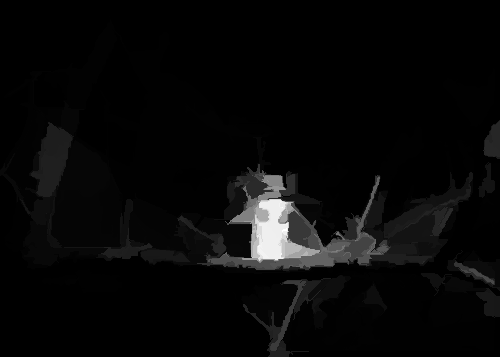} &
        \includegraphics[width=0.155\linewidth]{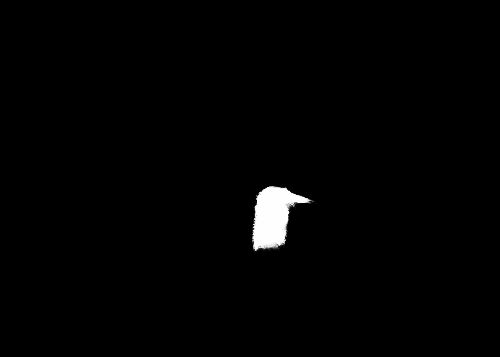} \\
        \includegraphics[width=0.155\linewidth]{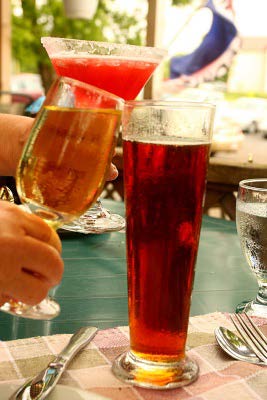}  &
        \includegraphics[width=0.155\linewidth]{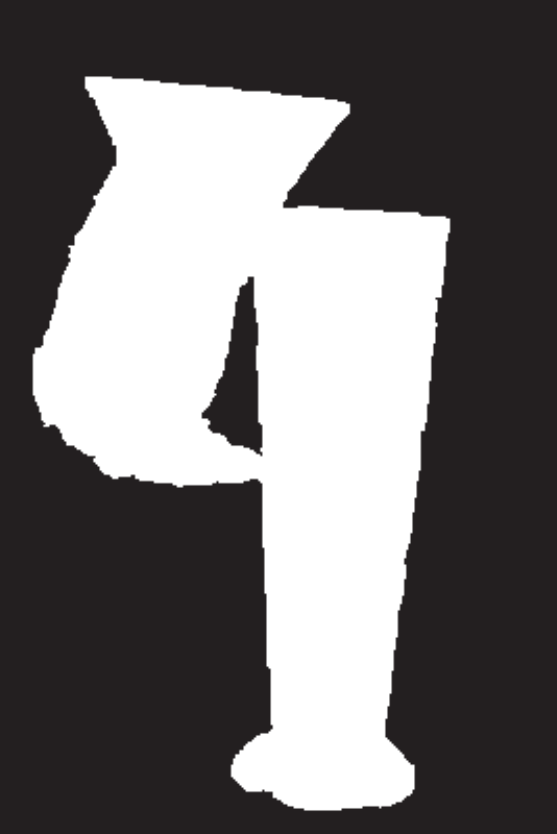} &
        \includegraphics[width=0.155\linewidth]{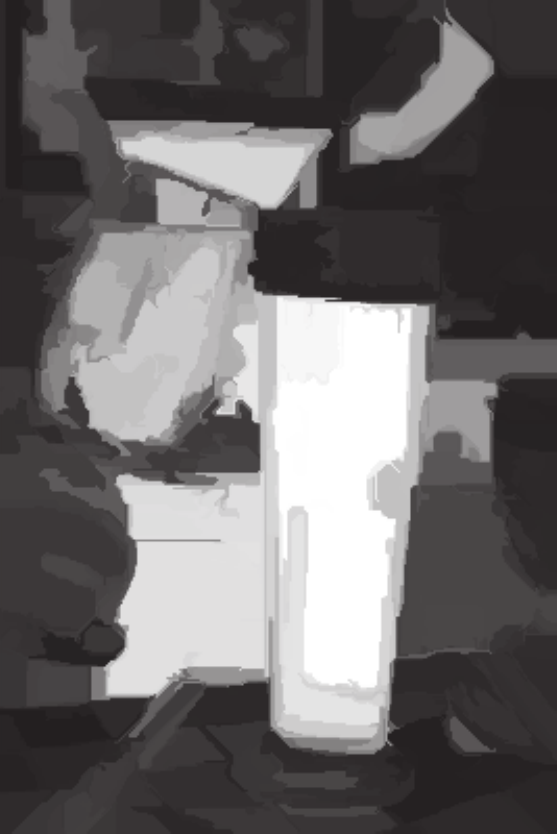} &
        \includegraphics[width=0.155\linewidth]{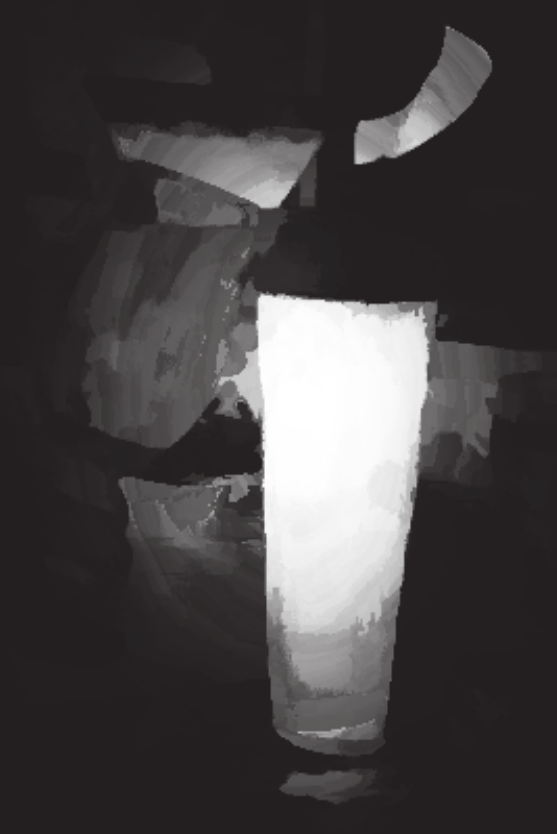} &
        \includegraphics[width=0.155\linewidth]{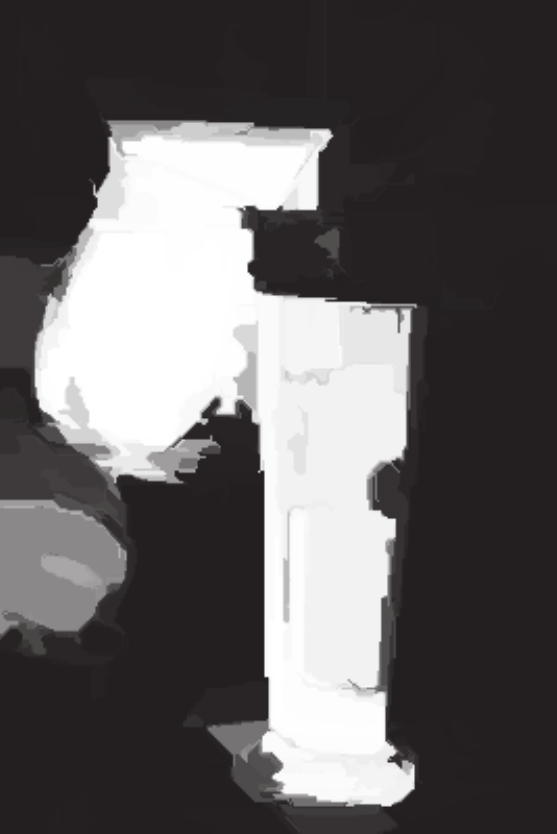} &
        \includegraphics[width=0.155\linewidth]{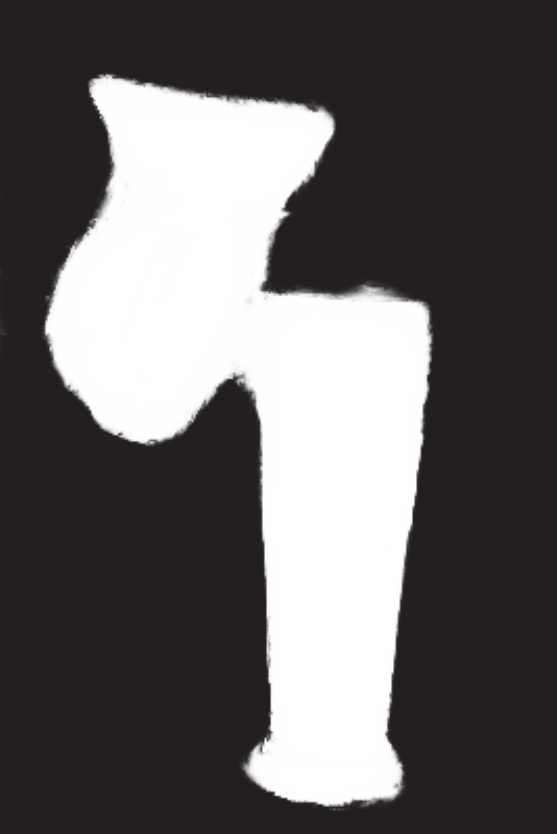} \\
        IMG & GT & DRFI \cite{jiang2013salient} & DSR \cite{li2013saliency} & MDF \cite{li2015visual}  & DSS \cite{hou2016deeply}
  \end{tabular}
  \caption{Visual comparisons of two best classic methods (DRFI and DSR), according to~\cite{borji2012salient} and two leading CNN-based methods (MDF and DSS).}
  \label{fig:vis_comp}
    \vspace{-10pt}
\end{figure}

Figure~\ref{fig:vis_comp} provides a visual comparison of maps generated by classic and CNN-based models.

\renewcommand{\AddImg}[3]{\subfigure[#1]{\includegraphics[#2]{#3}}}
\begin{figure*}[t!]
   \centering
   \AddImg{Content aware resizing \cite{ZhangC09} }{height=.115\linewidth,width=.23\linewidth}{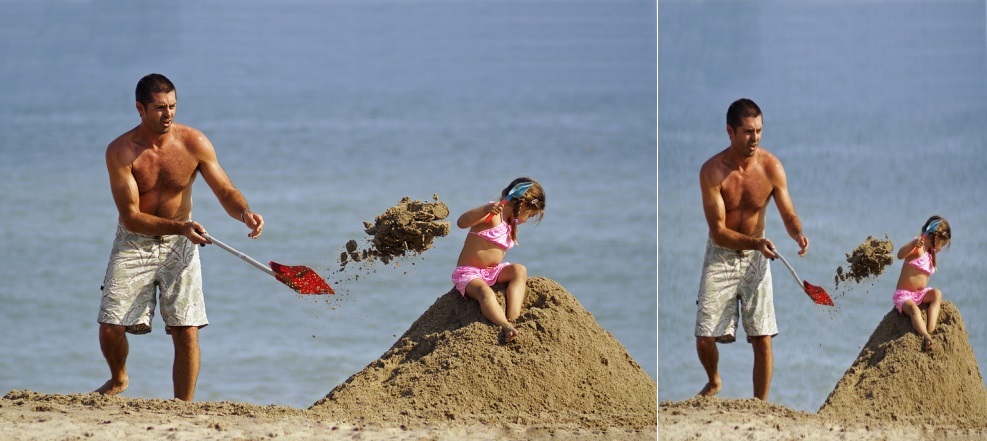} \hfill
   \AddImg{Image collage\cite{huang2011arcimboldo}}{height=.115\linewidth,width=.17\linewidth}{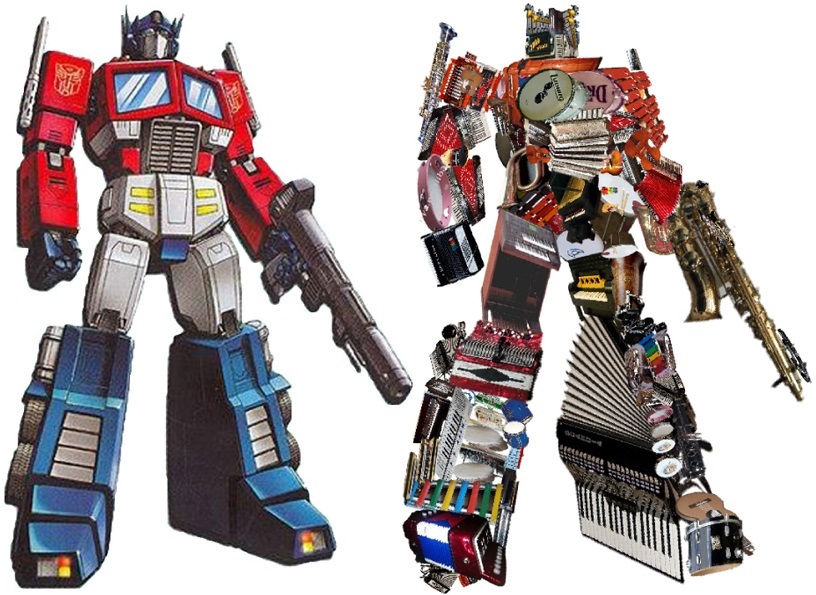}
   \AddImg{View selection \cite{liu2012web}}{height=.115\linewidth}{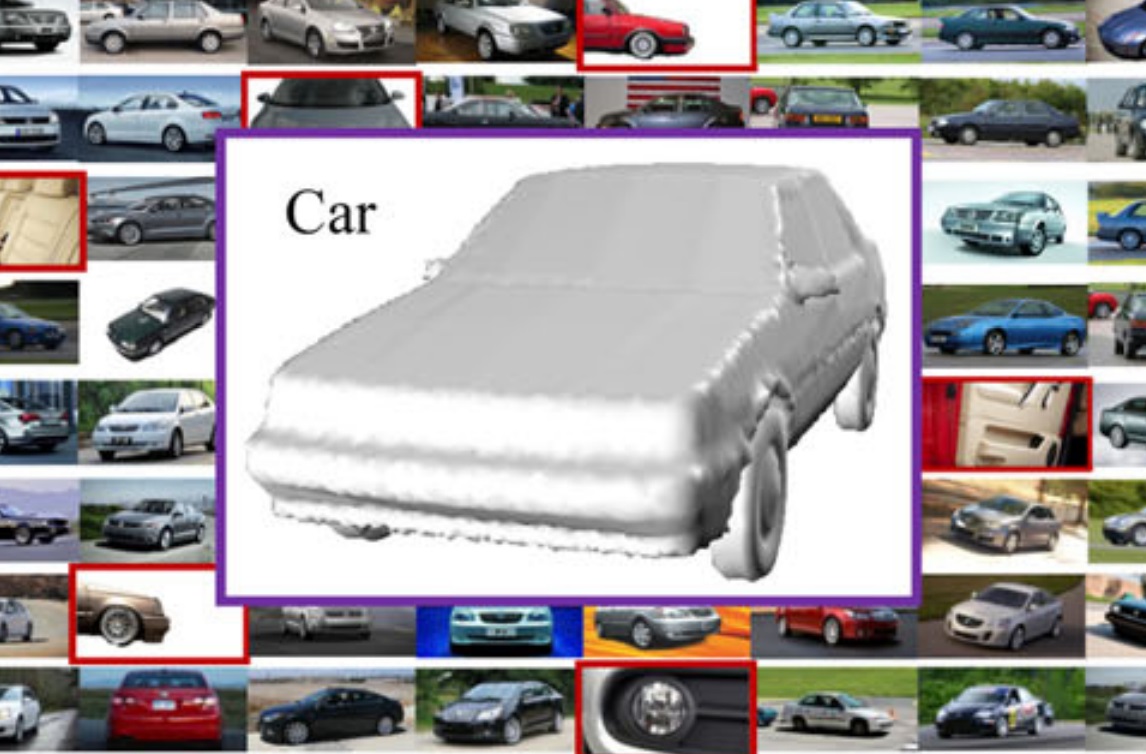}
   \AddImg{Unsupervised learning \cite{zhu2012unsupervised}}{height=.115\linewidth}{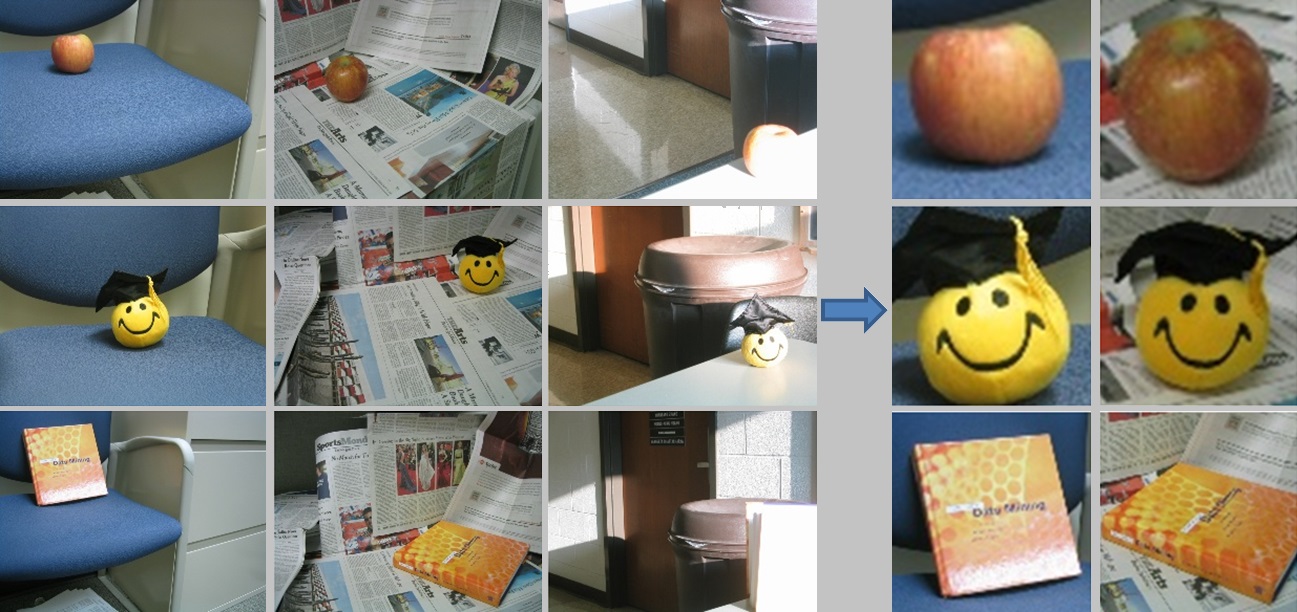}
   \AddImg{Mosaic \cite{margolin2013saliency}}{height=.115\linewidth}{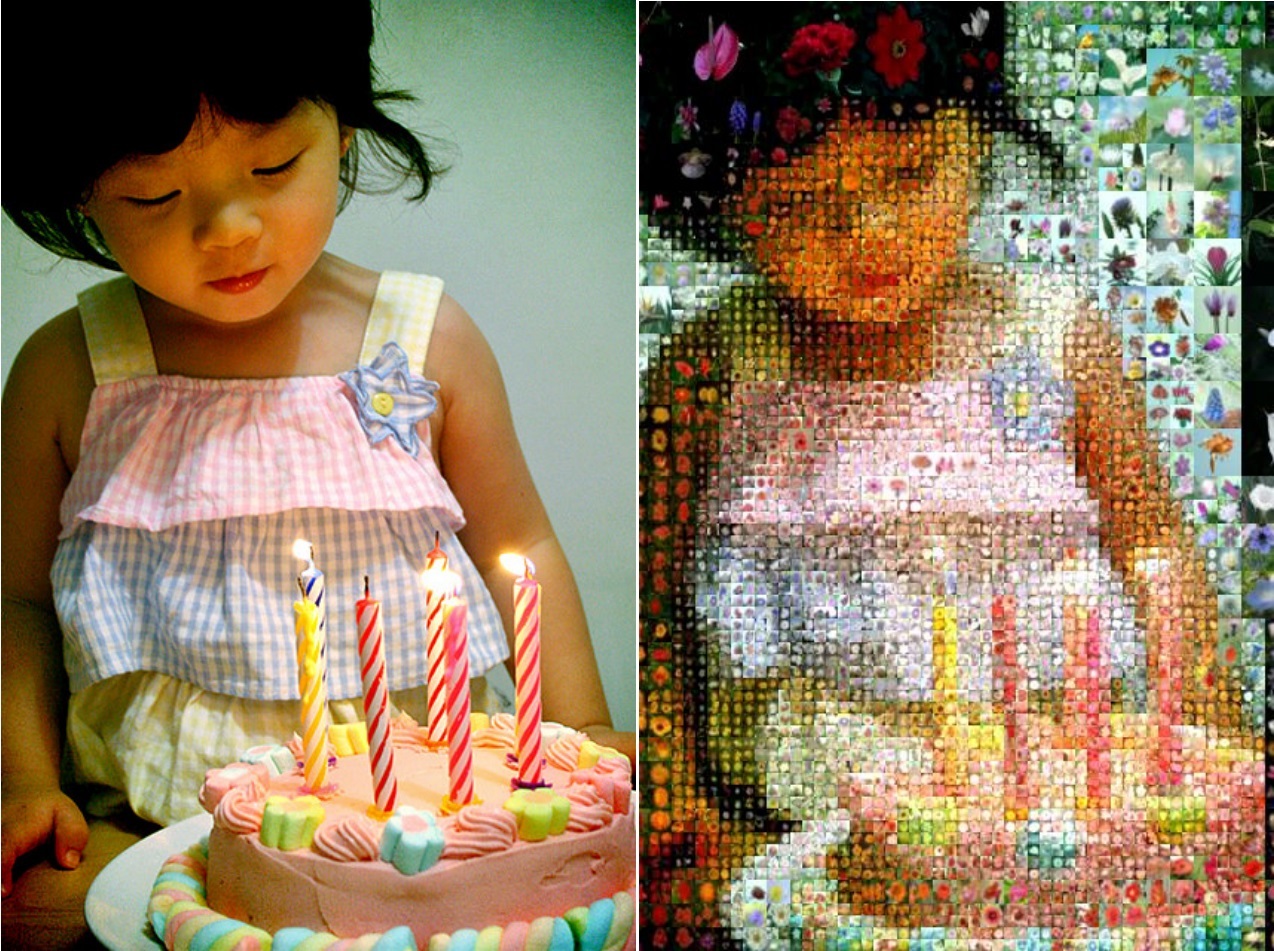}
   \\ \vspace{-.05in}
   \AddImg{Image montage \cite{chen2009sketch2photo}}{height=.1288\linewidth}{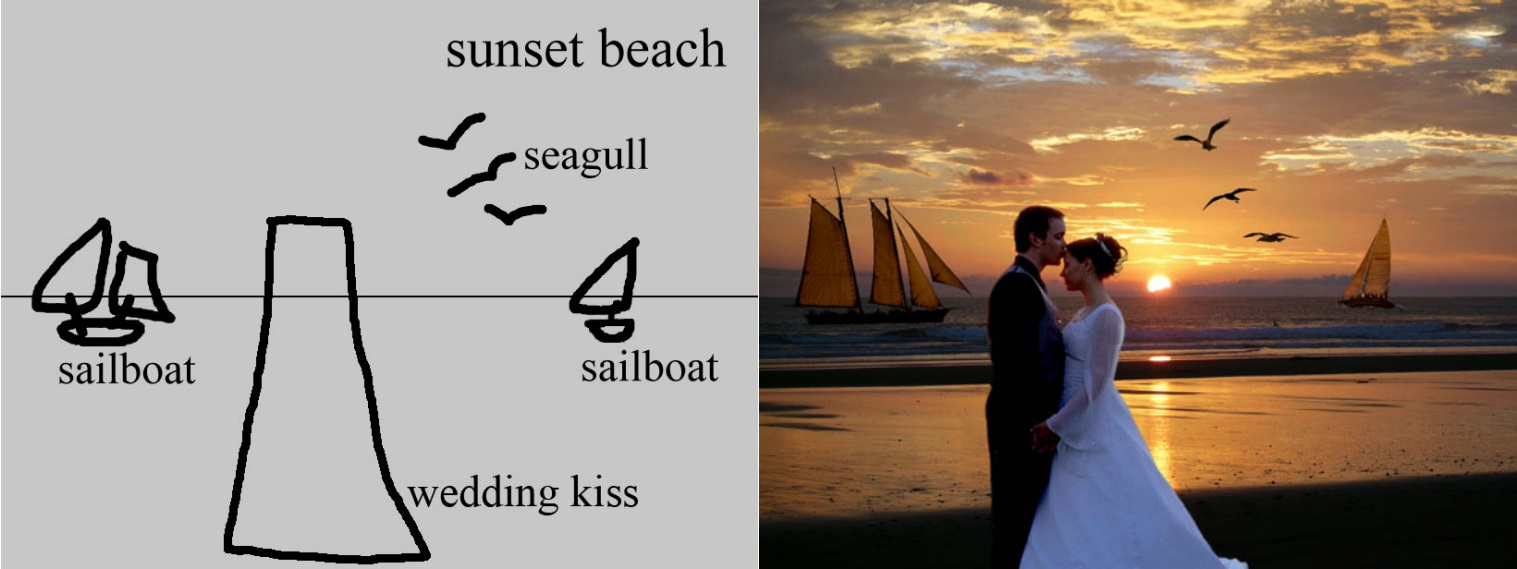}\hfill
   \AddImg{Object manipulation \cite{goldberg2012data}}{height=.1288\linewidth}{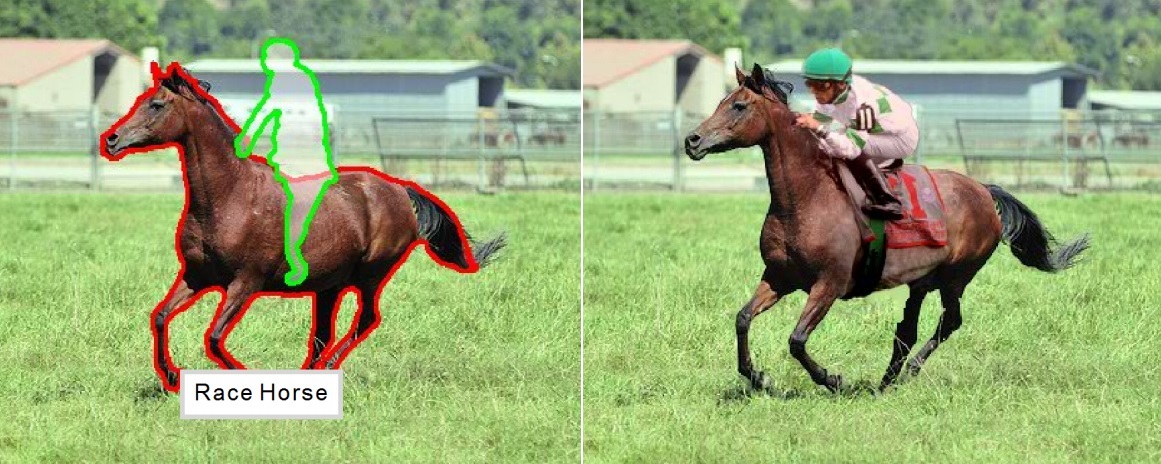}\hfill
   \AddImg{Semantic colorization \cite{chia2011semantic}}{height=.1288\linewidth}{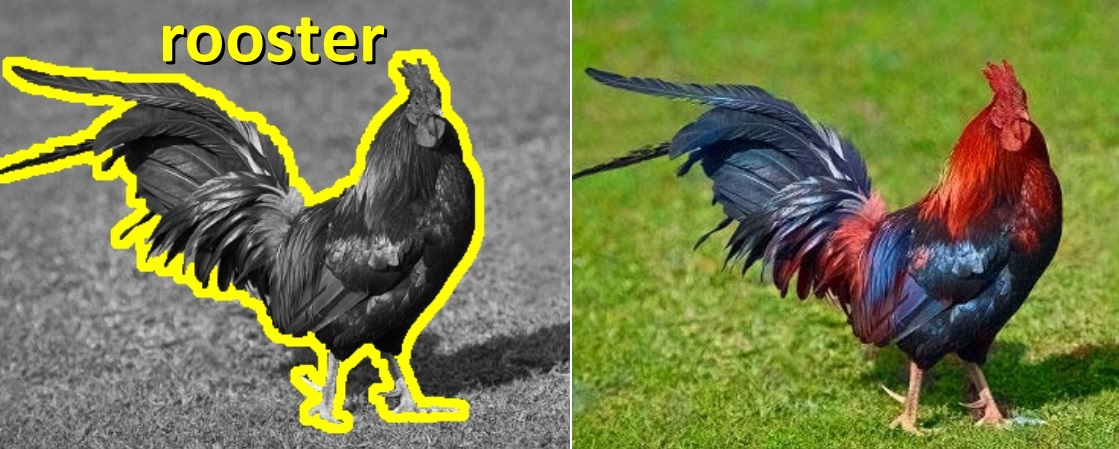}
   \\ \vspace{-.05in}
   \caption{Sample applications of salient object detection.
       Images are reproduced from corresponding references.
   }\label{fig:sampleApp}
   \vspace{-10pt}
\end{figure*}

\section{Applications of Salient Object Detection}
\label{sec:Applications}

The value of salient object detection models lies on their applications in many areas
of computer vision, graphics, and robotics.
Salient object detection models have been utilized for several applications such as object detection and recognition~
\cite{rutishauser2004bottom,kanan2010robust,moosmann2006learning,borji2011cost,borji2011scene,shen2013moving,ren2013region}, image and video compression~\cite{guo2010novel,itti2004automatic},
video summarization~\cite{ma2005generic,lee2012discovering,ji2012video},
photo collage/media re-targeting/cropping/thumb-nailing~\cite{goferman2010puzzle,wang2006picture,huang2011arcimboldo},
image quality assessment~\cite{ninassi2007does,liu2009studying,li2013color},
image segmentation~\cite{donoser2009saliency,li2011saliency,qin2013integration,johnson2010attention},
content-based image retrieval and image collection browsing \cite{chen2009sketch2photo,feng2010attention,sun2013image,li2013partial},
image editing and manipulating~\cite{chia2011semantic,liu2012web,margolin2013saliency,goldberg2012data},
visual tracking
\cite{stalder2013dynamic,li2013visual,garcia2012adaptive,borji2012adaptive,klein2010adaptive,frintrop2009most,zhang2010visual},
object discovery~\cite{karpathyobject,frintropcognitive},
and human-robot interaction~\cite{meger2008curious,sugano2010calibration}. 
\figref{fig:sampleApp} shows example applications.

\section{Datasets and Evaluation Measures}
\label{sec:Datasets}


\subsection{Salient Object Detection Datasets}

As more models have been proposed in the literature, more datasets have been introduced to further challenge saliency detection models.
Early attempts aim to collect images with salient objects being annotated with bounding boxes (\eg~\textbf{MSRA-A} and \textbf{MSRA-B}~\cite{LiuSZTS07Learn}), while later efforts annotate such salient objects with pixel-wise binary masks (\eg~\textbf{ASD}~\cite{achanta2009frequency} and \textbf{DUT-OMRON}~\cite{YangZLRY13Manifold}). Typically, images, which can be annotated with accurate masks, contain only limited objects (usually one) and simple background regions. On the contrary, recent attempts have been made to collect datasets with multiple objects in complex and cluttered backgrounds (\eg~\cite{borji2013stands,borjiTIP2014,liXiaodiCVPR2014}). As we mentioned in the Introduction section, a more sophisticated mechanism is required to determine the most salient object when several candidate objects are present in the same scene. For example, Borji~\cite{borjiTIP2014} and Li~\etal~\cite{liXiaodiCVPR2014} use the peak of human fixation map to determine which object is the most salient one (\ie~the one that humans look at the most; See section 1.2).

A list of 22 salient object datasets including 20 image datasets and 2 video datasets is shown in ~\tabref{tab:datasets}. Notice that all images or video frames in these datasets are annotated with binary masks or rectangles. Subjects are often asked to label the salient object in an image with one object (\eg~\cite{LiuSZTS07Learn}) or annotate the most salient one among several candidate objects (\eg~\cite{borji2013stands}). Some image datasets also provide the fixation data for each image collected during free-viewing task.

\subsection{Evaluation Measures}

Five universally-agreed, standard, and easy-to-compute measures for evaluating salient object detection models are described next.
For the sake of simplicity, we use $S$ to represent the predicted saliency map normalized to $[0,255]$ and $G$ be the ground-truth binary mask of salient objects. For a binary mask, we use $|\cdot|$ to represent the number of non-zero entries in the mask.

\myPara{Precision-recall (PR)} A saliency map $S$ is first converted to a binary mask $M$ and then $Precision$ and $Recall$ are computed by comparing $M$ with the ground-truth $G$:
\begin{align}
    Precision = \frac{|M\cap G|}{|M|}, \ \ Recall = \frac{|M\cap G|}{|G|}
\end{align}\label{eqn:precision_recall}
The binarization of $S$ is the key step in the evaluation. There are three popular ways to perform the binarization. In the first solution, Achanta~\etal~\cite{achanta2009frequency} propose the image-dependent adaptive threshold for binarizing $S$, which is computed as twice as the mean saliency of $S$:
\begin{equation}
    T_{a} = \frac{2}{W \times H} \sum_{x=1}^{W}\sum_{y=1}^{H} S(x,y),
\end{equation}
where $W$ and $H$ are the width and the height of the saliency map $S$, respectively.

The second way to binarize $S$ is to use a threshold that varies from 0 to 255. For each threshold, a pair of ($Precision$, $Recall$) scores are computed and used to plot a precision-recall (PR) curve. 

The third way to perform the binarization is to use the GrabCut-like algorithm (\eg~as in \cite{ChengPAMI}). Here, the PR curve is first computed and the threshold that leads to 95\% recall is selected. With this threshold, the initial binary mask is generated, which can be used to initialize the iterative GrabCut segmentation~\cite{RotherKB04Grab}. After several iterations, the binary mask can be gradually refined.

\myPara{F-measure} Often, neither $Precision$ nor $Recall$ can fully evaluate the quality of a saliency map.
To this end, the F-measure is proposed as the weighted harmonic mean of $Precision$ and $Recall$ with a non-negative weight $\beta^2$:
\begin{equation}
F_{\beta} = \frac{(1+\beta^{2}) Precision  \times  Recall}{\beta^{2} Precision + Recall}.
\label{eq:FMeasure}
\end{equation}
As suggested in many salient object detection works (\eg~\cite{achanta2009frequency}), $\beta^{2}$ is often set to $0.3$ to weigh $Precision$ more. The reason is because recall rate is not as important as precision (see also \cite{liu2011learning}). For instance, $100\%$ recall can be easily achieved by setting the whole map to be foreground.


\myPara{Receiver operating characteristics (ROC) curve} As above, false positive ($FPR$) and true positive rates ($TPR$) can be computed when binarizing the saliency map with a set of fixed thresholds:
\begin{align}
    TPR = \frac{|M\cap G|}{|G|}, \ \ FPR = \frac{|M\cap G|}{|M\cap G| + |\bar{M}\cap\bar{G}|}
\end{align}
where $\bar{M}$ and $\bar{G}$ denote the opposite of the binary mask $M$ and ground-truth $G$, respectively.
The ROC curve is the plot of $TPR$ versus $FPR$ by testing all possible thresholds.

\myPara{Arear under ROC curve (AUC)}
While ROC is a 2D representation of a model's performance, the AUC distills this information into a single scalar.
As the name implies, it is calculated as the area under the ROC curve. A perfect model will score an AUC of 1, while random guessing will score an AUC of around 0.5.

\myPara{Mean absolute error (MAE)}
The overlap-based evaluation measures introduced above do not consider the true negative saliency assignments,
\ie~the pixels correctly marked as non-salient. They favors methods that successfully assign high saliency to salient pixels but fail to detect non-salient regions. Moreover, for some applications~\cite{avidan2007seam}, the quality of the weighted continuous saliency maps may be of higher interest than the binary masks. For a more comprehensive comparison, it is recommended to evaluate the mean absolute error (MAE) between the continuous saliency map $S$ and the binary ground-truth $G$, both normalized in the range [0, 1]. The MAE score is defined as:
\begin{equation}
MAE = \frac{1}{W \times H} \sum_{x=1}^{W}\sum_{y=1}^{H} || S(x,y) - G(x,y)||.
\label{MAE}
\end{equation}

Please refer to~\cite{borji2015salient} for more details on datasets and scores\footnote{The code for evaluation measures is available at~\url{http://mmcheng.net/salobjbenchmark}.} in the filed of salient object detection.


\begin{figure}[t]
    \renewcommand{\arraystretch}{0.9}
    \renewcommand{\tabcolsep}{.7mm}
    \centering
    \small
    \begin{tabular}{|l||l|l|c|l|l|c|c|c|c|}\hline
     \textbf{Dataset} & \textbf{Year} & \textbf{Imgs}           & \textbf{Obj}  &  \textbf{Ann} &  \textbf{Resolution} & \textbf{Sbj}             & \textbf{Eye}   & \textbf{I/V}  \\
    \hline
    \hline
    \textbf{MSRA-A}  \cite{LiuSZTS07Learn,MSRAdb} & 2007 & 20K & $\sim$1 & BB & 400 $\times$ 300 & 3 & -  & I \\
    \textbf{MSRA-B}  \cite{LiuSZTS07Learn,MSRAdb} & 2007 &  5K & $\sim$1 & BB & 400 $\times$ 300 & 9 & -  & ,,\\
    \textbf{SED1}  \cite{alpert2007image,borji2012salient} & 2007 &100 & 1 & PW & $\sim$300 $\times$ 225 & 3 & -  &,,\\
    \textbf{SED2}  \cite{alpert2007image,borji2012salient} & 2007 &100 & 2 & PW & $\sim$300 $\times$ 225 & 3 & -  & ,,\\
     \textbf{ASD}  \cite{achanta2009frequency,LiuSZTS07Learn} & 2009 & 1000 & $\sim$1 & PW & 400 $\times$ 300 & 1 & -  &,,\\
    \textbf{SOD}  \cite{movahedi2010design,arbelaez2011contour} & 2010 &300 & $\sim$3 & PW & 481 $\times$ 321 & 7 & -  &,,\\
    \textbf{iCoSeg}  \cite{batra2010icoseg} & 2010 &643 & $\sim$1 & PW & $\sim$500 $\times$ 400 & 1 & -  &,,\\
    \textbf{MSRA5K}  \cite{jiang2011automatic,LiuSZTS07Learn} & 2011 &5K & $\sim$1 & PW & 400 $\times$ 300 & 1 & -  &,,\\
     \textbf{Infrared}  \cite{brown2011multi,wang2013multi} & 2011 &900 & $\sim$5 & PW & 1024 $\times$ 768 & 2 & 15  &,,\\
    \textbf{ImgSal}  \cite{li2013visual} & 2013 &235 & $\sim$ 2 & PW & 640 $\times$ 480 & 19 & 50  &,,\\
    \textbf{CSSD}  \cite{yan2013hierarchical} & 2013 &200 & $\sim$1 & PW & $\sim$400 $\times$ 300 & 1 & -  &,, \\
    \textbf{ECSSD}  \cite{yan2013hierarchical,ECSSDdb} & 2013  &1000 & $\sim$1 & PW & $\sim$400 $\times$ 300 & 1 & -  &,,\\
    \textbf{MSRA10K}   \cite{THUR15000db,LiuSZTS07Learn} & 2013 &10K & $\sim$1 & PW & 400 $\times$ 300 & 1 & -  & ,,\\
    \textbf{THUR15K}  \cite{THUR15000db,LiuSZTS07Learn} & 2013 &15K & $\sim$1 & PW & 400 $\times$ 300 & 1 & -  &,,\\
     \textbf{DUT-OMRON}  \cite{YangZLRY13Manifold} & 2013 &5,172 & $\sim$5 & BB & 400 $\times$ 400 & 5 & 5  &,,\\
    \textbf{Bruce-A}  \cite{borji2013stands,bruce2005saliency} &2013& 120 & $\sim$4 & PW & 681 $\times$ 511 & 70 & 20  &,,\\
     \textbf{Judd-A}  \cite{borjiTIP2014,JUDDdb} & 2014&900 & $\sim$5 & PW & 1024 $\times$ 768 & 2 & 15  & ,, \\
     \textbf{PASCAL-S}  \cite{liXiaodiCVPR2014} & 2014&850 & $\sim$5 & PW & variable & 12 & 8  &,,\\
     \textbf{UCSB}  \cite{koehler2014saliency} & 2014& 700 & $\sim$5 & PW & 405 $\times$ 405 & 100 & 8 &,,\\
     \textbf{OSIE}  \cite{xu2014predicting} & 2014& 700 & $\sim$5 & PW & 800 $\times$ 600 & 1 & 15 &,,\\
    \hline
    \hline
     \textbf{RSD}  \cite{li2009dataset} & 2009 & 62,356 & var. & BB & variable & 23 & -  & V\\
     \textbf{STC}  \cite{wu2011detection} & 2011 & 4,870 & $\sim$1 & BB & variable & 1 & - & ,, \\
    \hline
    \end{tabular}
    \caption{Overview of popular salient object datasets.
        Top: image datasets, Bottom: video datasets. Obj = objects per image; Ann = Annotation; Sbj = Subjects/Annotators; Eye = Eye tracking subjects; I/V = Image/Video.
    }\label{tab:datasets}
    \vspace{-10pt}
\end{figure}

\section{Discussions}\label{sec:discussion}

\subsection{Design Choices}

In the past two decades, hundreds of classic and deep learning based methods have been proposed for detecting and segmenting salient objects in scenes and a large number of design choices have been explored. Although great successes have been achieved recently, there is still a large room for improvement. Our detailed method summarization (see \figref{tab:salientObjModelsIntrin}
\& \figref{tab:SalModelExtrinsic})
does send some clear messages about the commonly used design choices,
which are valuable for the design of future algorithms. They are discussed next.
%
%

\subsubsection{Heuristic vs. Learning From Data}
Early methods were mainly based on heuristic (both local or global) cues to detect salient objects
\cite{achanta2009frequency,ChengPAMI,perazzi2012saliency,YangZLRY13Manifold}.
Recently, saliency models based on learning algorithms have shown to be very efficient (see~\tabref{tab:salientObjModelsIntrin} and \tabref{tab:SalModelExtrinsic}).
Among these models, deep learning based methods greatly outperform conventional heuristic methods because of their ability in learning large amount of extrinsic cues from large datasets.
%
%
Data-driven approaches for salient object detection seem to have a surprisingly good
generalization ability. An emerging question, however, is whether the data-driven ideas for salient object detection conflict with the ease of use of these models.
Most learning based approaches are only trained on a small subset of \textbf{MSRA5K} dataset,
and still consistently outperform other methods on all other datasets which have considerable differences. 
%
This suggests that it is worth to further explore
data-driven salient object detection without losing the simplicity
and ease-of-use advantages, in particular from an application point of view.

\subsubsection{Hand-crafted vs. CNN-based Features}

The first generation of learning-based methods were based on lots of hand-crafted features.
An obvious drawback of these methods is the generalization capability, especially when applied to complex cluttered scenes.
In addition, these methods mainly rely on over-segmentation algorithms, such as SLIC \cite{achanta2012slic}, yielding the incompleteness of most salient objects with high contrast components.
%
CNN-based models solve these problems, to some degree, even when complex scenes are considered.
Because of the ability of learning multi-level features, it is easy for CNNs to accurately locate where the salient objects are.
Low-level features such as edges enable sharpening boundaries of salient objects while high-level features allow incorporating semantic information to identify salient objects.

\subsubsection{Recent Advances in CNN-based Saliency Detection}

Various CNN-based architectures have been proposed recently.
Among these approaches, there are several promising choices that can be further explored in the future.
The first one regards models with deep supervision.
As shown in \cite{hou2016deeply}, deeply supervised networks strengthen the power of features at different layers.
The second choice is the encoder-decoder architecture, which has been adopted in many segmentation-related tasks.
These types of approaches gradually back-propagate high-level features to lower layers allowing effective fusion of multi-level features.
Another choice is exploiting stronger baseline models, such as using very deep ResNets \cite{He2016} instead of VGGNet \cite{simonyan2014very}.

\subsection{Dataset Bias}\label{sec:DatasetBais}

Datasets have been consequential in the rapid progress in saliency detection.
On the one hand, they supply large scale training data and enable comparing performance of competing algorithms.
On the other hand, each dataset is a unique sampling of an unlimitted
application domain, and contains a certain degree of bias.
%
%

To date, there seems to be a unanimous agreement on the presence of bias (i.e. skewness)
in underlying structures of datasets.
Consequently, some studies have addressed the effect of bias in image datasets.
For instance, Torralba \& Efros identify three biases in computer vision datasets,
namely: \textit{selection bias, capture bias} and \textit{negative set bias} \cite{torralba2011unbiased}.
Selection bias is caused by preference of a particular kind of image during data gathering.
It results in qualitatively similar images in a dataset.
This is witnessed by the strong color contrast (see \cite{liXiaodiCVPR2014,ChengPAMI})
in most frequently used salient object benchmark datasets~\cite{achanta2009frequency}.
Thus, two practices in dataset construction are preferred:
i) \textit{having independent image selection and annotation process} \cite{liXiaodiCVPR2014}, and
ii) \textit{detecting the most salient object first and then segmenting it.}
Negative set bias is the consequence of a lack of rich and unbiased negative set,
\ie~one should avoid concentrating on a particular image of interest
and datasets should represent the whole world.
Negative set bias may affect the ground-truth by incorporating annotator's personal
preference to some object types.
Thus, including a variety of images is encouraged in constructing a good dataset.
Capture bias conveys the effect of image composition on the dataset.
The most popular kind of such a bias is the tendency of composing objects in the central region of the image,
\ie~center bias.
The existence of bias in a dataset makes the quantitative comparisons very challenging
and sometimes even misleading. For instance, a trivial saliency model which consists of a Gaussian blob at the image center,
often scores higher than many fixation prediction models \cite{tatler2007central,judd2009learning,borji2013quantitative}.

\subsection{Future Directions}

Several promising research directions for constructing more effective models and benchmarks are discussed here.

\subsubsection{Beyond Working with Single Images}
Most benchmarks and saliency models discussed in this study deal with single images. Unfortunately, salient object detection on multiple input images, \eg~salient object detection on video sequences, co-salient object detection, and salient object detection over depth and light field images, are less explored.
One reason behind this is the limited availability of benchmark datasets on these problems.
For example, as mentioned in~\secref{sec:Datasets}, there are only two publicly available benchmark datasets for video saliency (mostly cartoons and news). For these videos, only bounding boxes are provided for the key frames to roughly localize salient objects.
Multi-modal data is becoming increasingly more accessible and affordable.
Integrating additional cues such as spatio-temporal consistency and depth will be beneficial for efficient salient object detection.

\subsubsection{Instance-Level Salient Object Detection}

Existing saliency models are object-agnostic (i.e., they do not split salient regions into objects). However, humans possess the capability of detecting salient objects at instance level.
Instance-level saliency can be useful in several applications, such as image editing and video compression.
%
%

Two possible approaches for instance-level saliency detection are as follows. The first one regards using an object detection or object proposal method, e.g., Fast-RCNN \cite{girshick2014rich}, to extract a stack of object bounding box candidates and then segment salient objects in them.
The second approach, initially proposed in \cite{li2017instance}, is leveraging edge information to distinguish different salient objects. 

\subsubsection{Versatile Network Architectures}
With the deeper understanding of researchers on CNNs, more and more interesting network architectures have been developed.
It has been shown that using advanced baseline models and network architectures~\cite{li2016deepX} can substantially improve the performance.
On the one hand, deeper networks do help better capture salient objects because of their ability in extracting high-level semantic information.
On the other hand, apart from high-level information, low-level features \cite{hou2016deeply,li2017instance} should also be considered to build high resolution saliency maps.

\subsubsection{Unanswered Questions}
Some remaining questions include: how many (salient) objects are necessarily to represent a scene? does map smoothing affect the scores and model ranking? how is salient object detection different from other fields? what is the best way to tackle the center bias in model evaluation? and what is the remaining gap between models and humans? A collaborative engagement with other related fields such as saliency for fixation prediction, scene labeling and categorization, semantic segmentation, object detection, and object recognition can help answer these questions, situate the field better, and identify future directions.

\vspace{-10pt}
\section{Summary and Conclusion}\label{sec:summary}

In this paper, we exhaustively review salient object detection literature with respect to its closely related areas.
Detecting and segmenting salient objects is very useful. Objects in images automatically capture more attention than background stuff, such as grass, trees and sky. Therefore, if we can detect salient or important objects first, then we can perform detailed reasoning and scene understanding at the next stage. Compared to traditional special-purpose object detectors, saliency models are general, typically fast, and
do not need heavy annotation. These properties allow processing a large number of images at low cost.

Exploring connections between salient object detection and fixation prediction models can help enhance performance of both types of models. In this regard, datasets that offer both salient object judgments of humans and eye movements are highly desirable. Conducting behavioral studies to understand how humans perceive and prioritize objects in scenes and how this concept is related to language, scene description and captioning, visual question answering, attributes, etc, can offer invaluable insights. Further, it is critical to focus more on evaluating and comparing salient object models to gauge future progress. Tackling dataset biases such as center bias and selection bias and moving towards more challenging images is important. 

Although salient object detection and segmentation methods have made great strides in recent years,
a very robust salient object detection algorithm that is able to generate high quality results for nearly all images is still missing. Even for humans, what is the most salient object in the image, is sometimes a quite ambiguous question. To this end, a general suggestion:\vspace{.1in}\\
\emph{Don't ask what segments can do for you, ask what you can do for the segments}\footnote{
\href{http://www.cs.berkeley.edu/~malik/student-tree-2010.pdf}
{http://www.cs.berkeley.edu/$\sim$malik/student-tree-2010.pdf}}. \hfill --- Jitendra Malik
\vspace{.1in}\\
is particularly important to build robust algorithms.
For instance, when dealing with noisy Internet images, although salient object detection and
segmentation methods do not guarantee robust performance on individual images,
their efficiency and simplicity makes it possible to automatically process a large
number of images. This allows filtering images for the purposes of reliability and accuracy,
running applications robustly
\cite{chen2009sketch2photo,ChengPAMI,ChengGroupSaliency,huang2011arcimboldo,liu2012web,chia2011semantic}
, and unsupervised learning \cite{zhu2012unsupervised}.

\bibliographystyle{CVM}

{\normalsize  \bibliography{egbib}}




\end{document}